\documentclass{article}

\usepackage{arxiv}

\usepackage{natbib}

\usepackage{algorithm}
\usepackage{algorithmic}
\usepackage{amsmath,graphicx}
\usepackage{amsfonts}
\usepackage{hyperref}
\usepackage{url}
\usepackage{bm}
\usepackage{adjustbox}
\usepackage{multirow,tabularx}

\usepackage{adjustbox}
\usepackage{booktabs}
\usepackage{multirow,subcaption,float}
\usepackage[table,xcdraw]{xcolor}

\newcommand\blfootnote[1]{%
  \begingroup
  \renewcommand\thefootnote{}\footnote{#1}%
  \addtocounter{footnote}{-1}%
  \endgroup
}

\title{A Supervised Contrastive Learning Pretrain-Finetune Approach for Time Series}

\author{
Trang H. Tran$^{1,2}$\thanks{Work done while an intern at IBM Research} \qquad 
Lam M. Nguyen$^{2}$ \qquad 
Kyongmin Yeo$^{2}$ \qquad
Nam Nguyen$^{2}$ \qquad 
Roman Vaculin$^{2}$ \\
$^{1}$ School of Operations Research and Information Engineering, Cornell University, Ithaca, NY, USA\\
$^{2}$ IBM Research, Thomas J. Watson Research Center, Yorktown Heights, NY, USA \\
\texttt{htt27@cornell.edu},
\texttt{LamNguyen.MLTD@ibm.com}
}

\begin{document}

\maketitle

\begin{abstract}
Foundation models have recently gained attention within the field of machine learning thanks to its efficiency in broad data processing. While researchers had attempted to extend this success to time series models, the main challenge is effectively extracting representations and transferring knowledge from pretraining datasets to the target finetuning dataset. To tackle this issue, we introduce a novel pretraining procedure that leverages supervised contrastive learning to distinguish features within each pretraining dataset. This pretraining phase enables a probabilistic similarity metric, which assesses the likelihood of a univariate sample being closely related to one of the pretraining datasets. Subsequently, using this similarity metric as a guide, we propose a fine-tuning procedure designed to enhance the accurate prediction of the target data by aligning it more closely with the learned dynamics of the pretraining datasets. Our experiments have shown promising results which demonstrate the efficacy of our approach.
\end{abstract}

\blfootnote{Correspondence to: Lam M. Nguyen.}

\section{Introductions}
\subsection{Motivations}
\paragraph{Foundation Models For Time Series.}
Lately, foundation models have gained significant prominence in the field of artificial intelligence and machine learning \citep{fms}, where notable examples of these models include BERT \citep{bert} and GPT-3 \citep{gpt3}. These models are characterized by their training on extensive datasets, typically employing self-supervised methods at a large scale, and they possess the capability to be adapted for a wide array of downstream tasks \citep{fms}. Therefore there have been efforts to extend this success to other applications, including time series \citep{onefits, rasul2023lagllama, xue2023make}.

\begin{figure}[h]
    \centering
    \begin{subfigure}{0.35\columnwidth}
    \centering
    \includegraphics[width=\textwidth]{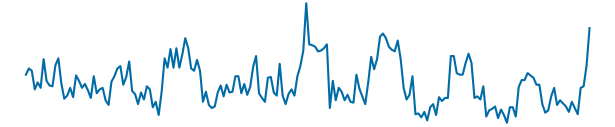}
    \includegraphics[width=\textwidth]{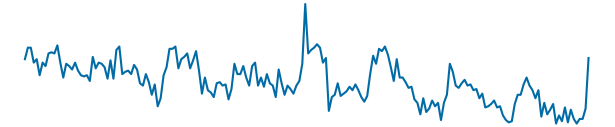}
    \includegraphics[width=\textwidth]{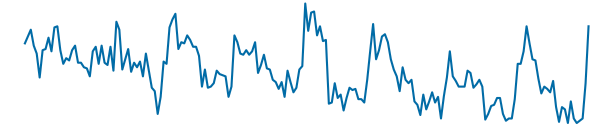}
    \includegraphics[width=\textwidth]{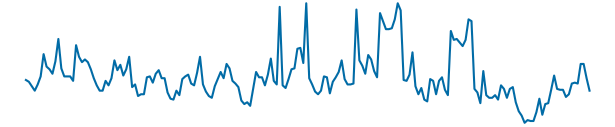}
    \includegraphics[width=\textwidth]{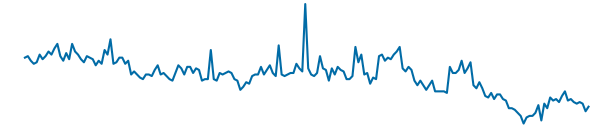}
    \caption{ETTh1 dataset} \label{fig:ETTh1}
    \end{subfigure}%
    \begin{subfigure}{0.35\columnwidth}
    \centering
    \includegraphics[width=\textwidth]{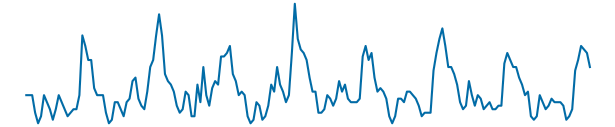}
    \includegraphics[width=\textwidth]{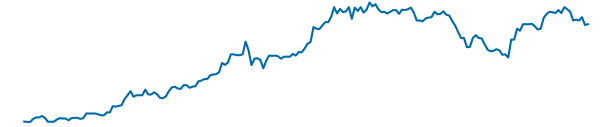}
    \includegraphics[width=\textwidth]{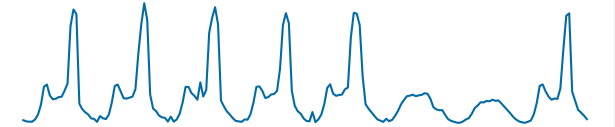}
    \includegraphics[width=\textwidth]{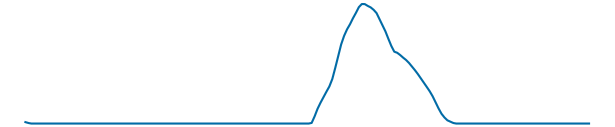}
    \includegraphics[width=\textwidth]{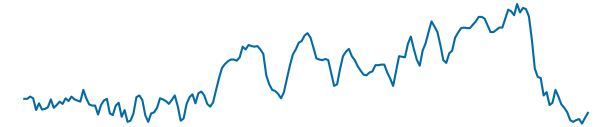}
    \caption{5 different datasets} \label{fig:diff}
    \end{subfigure}
    \caption{Plots of different features from ETTh1 and from five datasets. Description of datasets is in Section~\ref{subsec:exp_set}. Five different datasets featured are Electricity, Exchange-Rate, Traffic, Weather and ETTm1. 
    }
    \label{fig:dis}
\end{figure}

\paragraph{Challenges in General Representation Learning.}
One of the main challenges in training foundation models for time series is to address the discrepancy between pretraining and finetuning data \citep{Self-Supervised_Contrastive,yeh2023foundation}. 
As demonstrated in Figure \ref{fig:dis}, this discrepancy arises at various levels. In Figure \ref{fig:dis}(a), although the last feature has a slight deviation, most of the features bear some resemblance within the same datasets. However, for different datasets, the dynamics are vastly different, as shown in Figure \ref{fig:dis}(b). 
As a result, a foundation model should possess the capability to adapt with a heterogeneous collection of datasets. Therefore it is desirable to find a general representation to contain the diverse knowledge in the pretraining task \citep{Self-Supervised_Contrastive,zerveas2021transformer}. 

The next difficult task focuses on the design of the model to transfer the knowledge to finetuning task \citep{fawaz2018transfer}. It is reasonable to assume that the dynamic of the finetune datasets should be close to the dynamic of the collection of pretraining datasets in some sense. In this work, we consider the assumption that the representations of finetune dataset is within the span of the represenations of different pretraining datasets. 

Our approach is to differentiate the features that originate from different datasets by utilizing the learned representation of these features, therefore partially addressing the high-level discrepancy in time series dynamics and enhancing the knowledge within foundation model. We summarize our contributions below.


\paragraph{Contributions.}
\begin{itemize}
    \item We use a pretraining procedure with contrastive learning to differentiate the features in each pretraining dataset. This pretraining enables a probabilistic similarity metric to measure if a univariate sample is likely to share a similar dynamic with one of the pretraining datasets.
    \item Using the similarity metric, we propose a finetuning procedure to aid the correct prediction of the target data, by making the finetune representation closer to the learned representations of the corresponding pretrain datasets.
    \item Our experiments show that the pretrained models have promising performance compared to supervised training approaches. The finetuned model shows better generalization than prior approaches for some of the datasets, meanwhile having competitive results in other settings. 
\end{itemize}

\subsection{Related Work}

\paragraph{Time Series Forecasting}
There are two primary approaches for time series multi-step ahead predictions. Early approaches focuses on joint probability distributions of future system states by iteratively computing their evolution over time, typically using techniques like recurrent neural networks (RNNs) as demonstrated in \citep{rnn, Yeo22}. 
The second approach revolves around training a time series model capable of directly predicting future time steps based on historical data input. This includes multilayer perceptron (MLP)-based methods \citep{gardner1998artificial, lightts} along with convolutional neural networks \citep{oshea2015introduction, gu2018recent}. 
With the rise of attention-based models and their success in natural language processing \citep{attention}, attention-based time series models have gained popularity since they can discover the
temporal dependencies among time points. Nevertheless, these models face a challenge due to their quadratic time and memory complexity when dealing with learning long-range temporal correlations. To address this, LogTrans \citep{logtrans} and Pyraformer \citep{pyraformer} propose strategies to introduce sparsity bias and reduce computational complexity. Informer \citep{informer} and FEDformer \citep{fedformer} leverage the low-rank properties of the self-attention matrix to enhance performance.

In contrast, Autoformer \citep{autoformer} introduces a novel architecture with an auto-correlation mechanism as an alternative to traditional attention-based models. Conversely, in the work presented in \citep{linearmodels}, a different approach is taken with the use of a simple set of linear models and suggesting that these simplicity-driven models may outperform more complex structures. On the other hand, \citep{timesnet} learns the temporal patterns by exploring the multi-periodicity of time series and capture the temporal 2D-variations in 2D space.


\paragraph{Contrastive Learning.}

In recent years, there has been significant advancements in self-supervised representation learning \citep{ericsson2022self,jaiswal2020survey,misra2020self} with applications on time series \citep{cloc,tnc,t_loss,ts_tcc,ts2vec,sim_crl,btsf,tf_c,nguyen2023learning}. The common concept within these works is the idea of bringing an anchor and a positive sample closer in the embedding space, while separating the anchor and numerous negative samples. 
The work by \citep{supervised_contrastive} extends the self-supervised contrastive approach to the fully-supervised setting, enabling effective utilization of label information. 
The contribution of  \citep{supervised_contrastive} is considering multiple positive pairs per anchor, in addition to the numerous negative pairs, in contrast to self-supervised contrastive learning with a single positive pair. In this work, we utilize the supervised contrastive learning framework in \citep{supervised_contrastive} for the pretrain-finetune process. While there have been a self-supervised contrastive pretraining framework for time series \citep{Self-Supervised_Contrastive}, our approach is different since we ultilize the labels for training. 
\section{Problem Description}
\begin{figure*}[ht]
    \centering
    \begin{subfigure}{1\columnwidth}
    \centering
    \includegraphics[width=\textwidth]{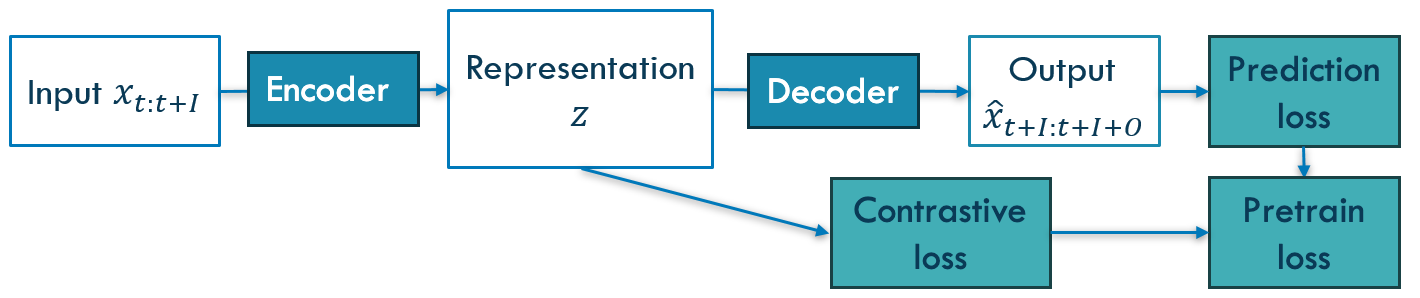}
    \caption{The encoder-decoder model and our pretrain loss. The pretrain loss has two component: a prediction loss on the model output, and a contrastive loss enforced on the representation $z$ of the model.} 
    \end{subfigure}%
    \hfill
    \begin{subfigure}{1\columnwidth}
    \centering
    \includegraphics[width=\textwidth]{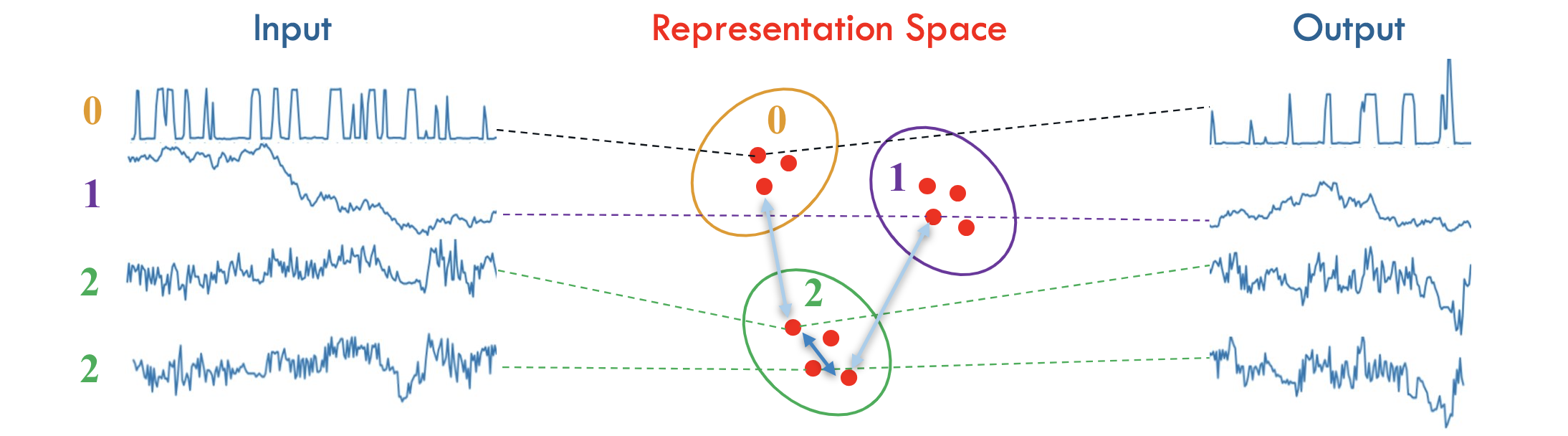}
    \caption{Each dataset is assigned a label and the contrastive loss maximizes the similarity (minimizes the difference) of the representations from the same label group while minimizes the similarity from different label groups. } 
    \end{subfigure}
    \caption{Description of our pretrain process with supervised contrastive learning}
    \label{fig:des}
\end{figure*}

We are given a collection of multivariate training datasets $\left\{X^k\right\}_{\text {pretrain}}, k = 1,\dots, P$, each one has sizes $T^k \times d^k$, where $T^k$ is the time dimension and $d^k$ is the number of features. The number of pretrain datasets is $P$. 
Our goal is to train a foundation model $M$ on the collection $\left\{X^k\right\}_{\text {pretrain}}$ and then finetune it to adapt with a new dataset $X_{\text{finetune}}$ of size $T^f\times d^f$ where $T^f$  and $d^f$ are the time dimension and the number of features of the finetune dataset, respectively.

We consider the time series forecasting problem where the model has the information of previous $I$ time steps and aims to predict the next $O$ future time steps.  Since the different datasets have different numbers of features, the designs of foundation model must have the capability to process such data. 
A typical approach to this problem is using channel independence, which learns a common model for every univariate time series data \citep{patchtst, han2023capacity, li2023revisiting, xue2023make}. 
It often consists of an encoder that transforms input data $x_{t: t+I}$ into a representation (context vector) and a decoder that generates the output sequence $x_{t+I: t+I+O}$ based on this context \citep{zhang2023multiscale, rasul2023lagllama}. The encoder and the decoder can have various structures ranging from simple fully connected layers to complex designs e.g. attention based models \citep{rasul2023lagllama, das2023longterm}. We describe this model structure in Figure \ref{fig:des}(a).

From the multivariate training datasets $\left\{X^k\right\}_{\text {pretrain}}$, we collects the univariate time series, which is further transformed into data samples using sliding windows. We note that the number of univariate data samples in each pretrain dataset is different. We build a pretrain sample collection which has equal number of data samples from each of the pretrain dataset $\left\{X^k\right\}_{\text {pretrain}}$.



\section{Pretrain-Finetune Approach with Supervised Contrastive Learning}

\subsection{Pretrain Process}
In this section, we describe our pretraining process. 
Our framework use a encoder-decoder model which takes the univariate time series as input. The pretrain loss function consists of two components where the first one is the mean squared error between the predicted values and the ground truth. The second component is a contrastive loss computed on the representation $z$ of the model. Accordingly: 
\begin{align}
    \text{Loss}_\text{pretrain}\left(x_{t: t+I}\right)= &\left\|\hat{x}_{t+I: t+I+O}-x_{t+I: t+I+O}\right\|^2 +\lambda \operatorname{SupCon}\left(x_{t: t+I}\right),
\end{align}
where
$\lambda$ is a regularizer and the (modified) supervised contrastive loss $\operatorname{SupCon}(x_{t: t+I})$ is
$$
\frac{-1}{|P(z)|} \sum_{p \in P(z)} \log \frac{\exp \left(z \cdot z_p / \tau\right)}{\sum_{n \in N(z)} \exp \left(z \cdot z_n / \tau\right)+\epsilon},
$$
where $z$ is the representation of the input data $x_{t: t+I}, \tau > 0$ is a scalar temperature parameter, $P(z)$ and $N(z)$ are the sets of positive and negative representations with $z$ in a batch of time series, respectively. The representations are negative if they comes from different datasets, and positive if they are from the same pretraining dataset. The operator $\cdot$ is a similarity metric, e.g. the inner dot product or cosine similarity. 

We apply this loss function for each data sample batch. By minimizing this contrastive loss, the model maximizes the similarity between $z$ with the set of positive representations and minimizes the similarity with the set of positive representations, within the batch. Compared to \citep{supervised_contrastive}, we make a modification of a small factor $\epsilon$ in the denominator since theoretically there could be no negative representation for a batch, still, the loss is well-defined.

\subsection{Probability of Similarity to A Pretrain Dataset}
After a pretraining process, the pretrained model $M_{\text {pretrained}}$ is more equipped with the ability to differentiate the heterogeneous temporal dynamics of time series data. 
However, the next question is how to leverage this knowledge in the finetune process i.e. how the model recognizes the dynamics it has learned in the past. In this section, we propose to use a quantity that approximate the probability of similarity to a prior-exposed pretrain dataset. This helps to analyze the model better and aids the finetuning process. 

Let $z$ be a representation of a finetune data sample and $\{z_l\}_{l= 1, 2, \dots }$ be all the representations of the pretraining samples, produced by the pretrained model $M_{\text {pretrained}}$. We note that those representations depends on the current learning model. We recall that $P$ is the number of pretraining datasets and the similarity metric is $z \cdot z_k$. Thus the approximate probability that the finetune data corresponding with $z$ comes from dataset $i$ is:
\begin{align}\label{eq:prob}
    p_i=\frac{\sum_{l \in \operatorname{Dataset}(i)} \exp \left(z \cdot z_l / \tau\right)}{\sum_{j=1, \ldots, C} \sum_{l \in \operatorname{Dataset}(j)} \exp \left(z \cdot z_l / \tau\right)}
\end{align}

This estimation naturally arises from the design of the supervised contrastive loss. Since the model maximizes $\exp \left(z \cdot z_p / \tau\right)$ where $z_p$ is the positive representation and minimizes $\exp \left(z \cdot z_n / \tau\right)$ where $z_n$ is the negative representation, then for a new representation $z$, the datasets which has similar dynamic to $z$ should have higher value of $\sum_{l \in \operatorname{Dataset}(i)} \exp \left(z \cdot z_l / \tau\right)$. Dividing this quantity by the sum for all the datasets, we get the estimated probability in \eqref{eq:prob}.

\subsection{Finetune Using Similarity Metrics}
\begin{figure}[hpt!]
    \centering
    \includegraphics[width = 0.4\textwidth]{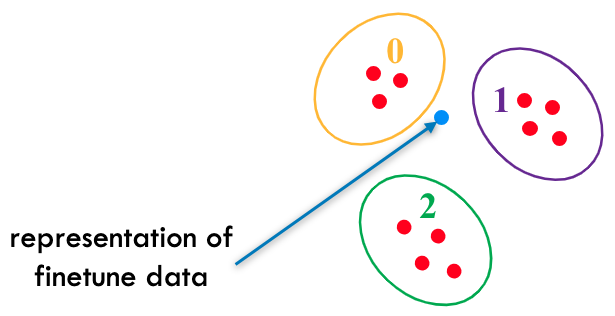}
    \caption{The representation of the finetune data can be close to some sample groups (0) and (1) more than other group (2). In such cases, we give priority to group (0) and then (1) and avoid being close to (2).
    }
    \label{fig:finetune}
\end{figure}

An estimated probability is a good tool to analyze the finetune sample data and gain insights on whether the finetune data is likely to belong to or behave similarly to any of the pretraining datasets. In this section, we propose to utilize this insight. Intuitively, if there is a high chance that a finetune data belongs to a pretrain dataset $i$, then it is best to for the model to use the dynamics learned by dataset $i$. On the other hand, if there is more than one dominant dataset e.g. $[0.4, 0.4, 0.1, 0.1]$, then it is beneficial to consider the datasets with high chances (see Figure \ref{fig:finetune}). This observation motivates our finetune process. 

The finetune loss function is similar to the pretrain loss that is consisting of a prediction component and a supervised contrastive component. Therefore:
\begin{align}
    \operatorname{Loss}_\text{finetune}\left(x_{t: t+I}\right)&=\left\|\hat{x}_{t+I: t+I+O}-x_{t+I: t+I+O}\right\|^2 +\lambda^{\prime} \mathrm{FTCon}\left(x_{t: t+I}\right),
\end{align}
where 
$\lambda^{\prime}$ is a regularizer and the finetune contrastive loss $\operatorname{FTCon}(x)$ is
$$
\frac{-1}{|P(z)|} \sum_{p \in P(z)} \log \frac{\exp \left(z \cdot z_p / \tau\right)}{\sum_{n \in N(z)} \exp \left(z \cdot z_n / \tau\right)+\epsilon},
$$
where $z$ is the (current) representation of the finetune input data $x_{t: t+I}$, the sets $P(z)$ and $N(z)$ of positive and negative representations of pretrain samples are defined as:
$$
\left\{\begin{array}{l}
i \in P(z) \text { if } p_i > 1 / P \\
i \in N(z) \text { if } p_i < 1 / P
\end{array}\right.
$$
where $p_i$ is the estimated probability in \eqref{eq:prob}.
The choice of $1/P$ stems from the fact that there is $P$ pretrain datasets i.e. higher probability than $1/P$ indicates higher similarity, which is considered positive representations. If a model predict $p_i = 1/P$, then it offers no information whether the finetune data is similar to dataset $i$ or not, and then dataset $i$ is discarded (not considered in the process). We note that $p_i$ is not fixed throughout the finetuning process, as it changes with the weights of the model as the representations is updated. 

The advantage of the finetune loss is two-fold. On the one hand, the information of $p_i$ helps the model to find better representations of the finetune data that are closer to that of the pretraining datasets to which it is similar. On the other hand, when the representations learned by the pretrain model is not good enough for the finetune data (and might lead to inaccurate/ mismatch probability) then the prediction loss helps to find better representations which gradually change their estimated probability, and give better context for the finetune time series dynamic.

\section{Experiments}
\subsection{Experiment Settings}\label{subsec:exp_set}
\paragraph{Datasets}
To maintain fair comparison between our work and prior benchmark, we test our model using the standard experiment procedure as in \citep{autoformer}.
Our experiments use following real-world datasets: ETT, Traffic, Electricity, Weather, Exchange-Rate and ILI  \citep{autoformer}. 
The ETT dataset contains information collected from electricity transformers, load and oil temperature data recorded at 15-minute intervals spanning from July 2016 to July 2018. The Electricity dataset records the hourly electricity consumption of 321 customers over a three-year period. Exchange-Rate dataset contains daily exchange rate data from eight different countries spanning from 1990 to 2016. The Traffic dataset records hourly data from the California Department of Transportation, including road occupancy rates measured by various sensors throughout the San Francisco Bay area. The Weather dataset provides 21 meteorological indicators, with data recorded at 10-minute intervals throughout the year 2020. Lastly, the ILI includes the influenza-like illness (ILI) patients data from CDC every week between 2002 and 2021. 

We follow the standard experiment procedure as in \citep{autoformer}. The time series data is split into training, validation and test sets in chronological order by the ratio of 7:1:2 for all the data sets. To ensure fair comparison, in our pretraining process we only use the training proportions of the original datasets. In our test, we use the same metrics as the prior reference \citep{autoformer} with batch size 32. 


\setlength{\tabcolsep}{1pt}
\renewcommand{\arraystretch}{1.14}

\begin{table*}[ht!]
\caption{Comparisons of the test performance from our pretrained model and other supervised learning models*.\\}
\label{tab:pretrain}
\begin{adjustbox}{width=0.99\textwidth}
\begin{tabular}{|c|c|cc|cc|cc|cc|cc|cc|cc|cc|cc|}
\hline
 & Models & \multicolumn{2}{c|}{Our PT model} & \multicolumn{2}{c|}{Ratio} & \multicolumn{2}{c|}{TimesNet} & \multicolumn{2}{c|}{ETSformer} & \multicolumn{2}{c|}{LightTS} & \multicolumn{2}{c|}{DLinear} & \multicolumn{2}{c|}{FEDformer} & \multicolumn{2}{c|}{Stationary} & \multicolumn{2}{c|}{Autoformer} \\
\multirow{-2}{*}{Data} & Metric & MSE & MAE & MSE & MAE & MSE & MAE & MSE & MAE & MSE & MAE & MSE & MAE & MSE & MAE & MSE & MAE & MSE & MAE \\
\hline
\cellcolor[HTML]{FFF7E7} & 96 & 0.415 & 0.430 & \cellcolor[HTML]{CCDDF8}1.228 & \cellcolor[HTML]{DFEAFB}1.147 & {\color[HTML]{FF0000} \textbf{0.338}} & {\color[HTML]{0000FF} \underline{ 0.375}} & 0.375 & 0.398 & 0.374 & 0.400 & {\color[HTML]{0000FF} \underline{ 0.345}} & {\color[HTML]{FF0000} \textbf{0.372}} & 0.379 & 0.419 & 0.386 & 0.398 & 0.505 & 0.475 \\
\cellcolor[HTML]{FFF7E7} & 192 & 0.471 & 0.451 & \cellcolor[HTML]{C5D9F8}1.259 & \cellcolor[HTML]{DAE7FA}1.165 & {\color[HTML]{FF0000} \textbf{0.374}} & {\color[HTML]{FF0000} \textbf{0.387}} & 0.408 & 0.410 & 0.400 & 0.407 & {\color[HTML]{0000FF} \underline{ 0.380}} & {\color[HTML]{0000FF} \underline{ 0.389}} & 0.426 & 0.441 & 0.459 & 0.444 & 0.553 & 0.496 \\
\cellcolor[HTML]{FFF7E7} & 336 & 0.513 & 0.479 & \cellcolor[HTML]{C7DAF8}1.251 & \cellcolor[HTML]{DAE7FA}1.165 & {\color[HTML]{FF0000} \textbf{0.410}} & {\color[HTML]{FF0000} \textbf{0.411}} & 0.435 & 0.428 & 0.438 & 0.438 & {\color[HTML]{0000FF} \underline{ 0.413}} & {\color[HTML]{0000FF} \underline{ 0.413}} & 0.445 & 0.459 & 0.495 & 0.464 & 0.621 & 0.537 \\
\multirow{-4}{*}{\cellcolor[HTML]{FFF7E7}\rotatebox{90}{ETTm1}} & 720 & 0.565 & 0.521 & \cellcolor[HTML]{D7E4FA}1.182 & \cellcolor[HTML]{DCE8FB}1.158 & {\color[HTML]{0000FF} \underline{ 0.478}} & {\color[HTML]{FF0000} \textbf{0.450}} & 0.499 & 0.462 & 0.527 & 0.502 & {\color[HTML]{FF0000} \textbf{0.474}} & {\color[HTML]{0000FF} \underline{ 0.453}} & 0.543 & 0.490 & 0.585 & 0.516 & 0.671 & 0.561 \\
\hline
\cellcolor[HTML]{FFF7E7} & 96 & 0.413 & 0.417 & \cellcolor[HTML]{EFF4FD}1.076 & \cellcolor[HTML]{F7FAFE}1.037 & {\color[HTML]{0000FF} \underline{ 0.384}} & {\color[HTML]{0000FF} \underline{ 0.402}} & 0.494 & 0.479 & 0.424 & 0.432 & 0.386 & {\color[HTML]{FF0000} \textbf{0.400}} & {\color[HTML]{FF0000} \textbf{0.376}} & 0.419 & 0.513 & 0.491 & 0.449 & 0.459 \\
\cellcolor[HTML]{FFF7E7} & 192 & 0.455 & 0.442 & \cellcolor[HTML]{F6F9FE}1.044 & \cellcolor[HTML]{F9FBFF}1.030 & {\color[HTML]{0000FF} \underline{ 0.436}} & {\color[HTML]{FF0000} \textbf{0.429}} & 0.538 & 0.504 & 0.475 & 0.462 & 0.437 & {\color[HTML]{0000FF} \underline{ 0.432}} & {\color[HTML]{FF0000} \textbf{0.420}} & 0.448 & 0.534 & 0.504 & 0.500 & 0.482 \\
\cellcolor[HTML]{FFF7E7} & 336 & 0.496 & 0.467 & \cellcolor[HTML]{FDFEFF}1.010 & \cellcolor[HTML]{F4CCCC}0.996 & 0.491 & 0.469 & 0.574 & 0.521 & 0.518 & 0.488 & {\color[HTML]{0000FF} \underline{ 0.481}} & {\color[HTML]{FF0000} \textbf{0.459}} & {\color[HTML]{FF0000} \textbf{0.459}} & {\color[HTML]{0000FF} \underline{ 0.465}} & 0.588 & 0.535 & 0.521 & 0.496 \\
\multirow{-4}{*}{\cellcolor[HTML]{FFF7E7}\rotatebox{90}{ETTh1}} & 720 & 0.537 & 0.525 & \cellcolor[HTML]{F9FBFF}1.031 & \cellcolor[HTML]{F4F8FE}1.050 & 0.521 & {\color[HTML]{FF0000} \textbf{0.500}} & 0.562 & 0.535 & 0.547 & 0.533 & 0.519 & 0.516 & {\color[HTML]{FF0000} \textbf{0.506}} & {\color[HTML]{0000FF} \underline{ 0.507}} & 0.643 & 0.616 & {\color[HTML]{0000FF} \underline{ 0.514}} & 0.512 \\
\hline
\cellcolor[HTML]{FFF7E7} & 96 & 0.103 & 0.239 & \cellcolor[HTML]{F4CCCC}0.963 & \cellcolor[HTML]{FBFCFF}1.021 & 0.107 & 0.234 & {\color[HTML]{FF0000} \textbf{0.085}} & {\color[HTML]{FF0000} \textbf{0.204}} & 0.116 & 0.262 & {\color[HTML]{0000FF} \underline{ 0.088}} & {\color[HTML]{0000FF} \underline{ 0.218}} & 0.148 & 0.278 & 0.111 & 0.237 & 0.197 & 0.323 \\
\cellcolor[HTML]{FFF7E7} & 192 & 0.184 & 0.325 & \cellcolor[HTML]{F4CCCC}0.814 & \cellcolor[HTML]{F4CCCC}0.945 & 0.226 & 0.344 & {\color[HTML]{0000FF} \underline{ 0.182}} & {\color[HTML]{FF0000} \textbf{0.303}} & 0.215 & 0.359 & {\color[HTML]{FF0000} \textbf{0.176}} & {\color[HTML]{0000FF} \underline{ 0.315}} & 0.271 & 0.380 & 0.219 & 0.335 & 0.300 & 0.369 \\
\cellcolor[HTML]{FFF7E7} & 336 & {\color[HTML]{FF0000} \textbf{0.296}} & {\color[HTML]{FF0000} \textbf{0.420}} & \cellcolor[HTML]{F4CCCC}0.807 & \cellcolor[HTML]{F4CCCC}0.938 & 0.367 & 0.448 & 0.348 & 0.428 & 0.377 & 0.466 & {\color[HTML]{0000FF} \underline{ 0.313}} & {\color[HTML]{0000FF} \underline{ 0.427}} & 0.460 & 0.500 & 0.421 & 0.476 & 0.509 & 0.524 \\
\multirow{-4}{*}{\cellcolor[HTML]{FFF7E7}\rotatebox{90}{Exchange}} & 720 & {\color[HTML]{FF0000} \textbf{0.537}} & {\color[HTML]{FF0000} \textbf{0.588}} & \cellcolor[HTML]{F4CCCC}0.557 & \cellcolor[HTML]{F4CCCC}0.788 & 0.964 & 0.746 & 1.025 & 0.774 & {\color[HTML]{0000FF} \underline{ 0.831}} & 0.699 & 0.839 & {\color[HTML]{0000FF} \underline{ 0.695}} & 1.195 & 0.841 & 1.092 & 0.769 & 1.447 & 0.941 \\
\hline
\cellcolor[HTML]{FFF7E7} & 96 & 0.253 & 0.336 & \cellcolor[HTML]{8EB4F0}1.506 & \cellcolor[HTML]{CBDCF8}1.235 & {\color[HTML]{FF0000} \textbf{0.168}} & {\color[HTML]{FF0000} \textbf{0.272}} & 0.187 & 0.304 & 0.207 & 0.307 & 0.197 & 0.282 & 0.193 & 0.308 & {\color[HTML]{0000FF} \underline{ 0.169}} & {\color[HTML]{0000FF} \underline{ 0.273}} & 0.201 & 0.317 \\
\cellcolor[HTML]{FFF7E7} & 192 & 0.247 & 0.337 & \cellcolor[HTML]{B3CCF5}1.342 & \cellcolor[HTML]{DAE7FA}1.166 & {\color[HTML]{0000FF} \underline{ 0.184}} & 0.289 & 0.199 & 0.315 & 0.213 & 0.316 & 0.196 & {\color[HTML]{FF0000} \textbf{0.285}} & 0.201 & 0.315 & {\color[HTML]{FF0000} \textbf{0.182}} & {\color[HTML]{0000FF} \underline{ 0.286}} & 0.222 & 0.334 \\
\cellcolor[HTML]{FFF7E7} & 336 & 0.268 & 0.360 & \cellcolor[HTML]{B0CBF5}1.354 & \cellcolor[HTML]{D3E2F9}1.200 & {\color[HTML]{FF0000} \textbf{0.198}} & {\color[HTML]{FF0000} \textbf{0.300}} & 0.212 & 0.329 & 0.230 & 0.333 & 0.209 & {\color[HTML]{0000FF} \underline{ 0.301}} & 0.214 & 0.329 & {\color[HTML]{0000FF} \underline{ 0.200}} & 0.304 & 0.231 & 0.338 \\
\multirow{-4}{*}{\cellcolor[HTML]{FFF7E7}\rotatebox{90}{Electricity}} & 720 & 0.310 & 0.398 & \cellcolor[HTML]{A4C2F3}1.409 & \cellcolor[HTML]{C9DBF8}1.244 & {\color[HTML]{FF0000} \textbf{0.220}} & {\color[HTML]{FF0000} \textbf{0.320}} & 0.233 & 0.345 & 0.265 & 0.360 & 0.245 & 0.333 & 0.246 & 0.355 & {\color[HTML]{0000FF} \underline{ 0.222}} & {\color[HTML]{0000FF} \underline{ 0.321}} & 0.254 & 0.361 \\
\hline
\cellcolor[HTML]{F1E5FF} & 96 & 0.200 & 0.296 & \cellcolor[HTML]{F0F5FD}1.070 & \cellcolor[HTML]{E7EFFC}1.109 & {\color[HTML]{FF0000} \textbf{0.187}} & {\color[HTML]{FF0000} \textbf{0.267}} & {\color[HTML]{0000FF} \underline{ 0.189}} & 0.280 & 0.209 & 0.308 & 0.193 & 0.292 & 0.203 & 0.287 & 0.192 & {\color[HTML]{0000FF} \underline{ 0.274}} & 0.255 & 0.339 \\
\cellcolor[HTML]{F1E5FF} & 192 & 0.286 & 0.360 & \cellcolor[HTML]{DEE9FB}1.149 & \cellcolor[HTML]{DAE7FA}1.165 & {\color[HTML]{FF0000} \textbf{0.249}} & {\color[HTML]{FF0000} \textbf{0.309}} & {\color[HTML]{0000FF} \underline{ 0.253}} & {\color[HTML]{0000FF} \underline{ 0.319}} & 0.311 & 0.382 & 0.284 & 0.362 & 0.269 & 0.328 & 0.280 & 0.339 & 0.281 & 0.340 \\
\cellcolor[HTML]{F1E5FF} & 336 & 0.424 & 0.452 & \cellcolor[HTML]{B7D0F6}1.321 & \cellcolor[HTML]{BFD5F7}1.288 & {\color[HTML]{0000FF} \underline{ 0.321}} & {\color[HTML]{FF0000} \textbf{0.351}} & {\color[HTML]{FF0000} \textbf{0.314}} & {\color[HTML]{0000FF} \underline{ 0.357}} & 0.442 & 0.466 & 0.369 & 0.427 & 0.325 & 0.366 & 0.334 & 0.361 & 0.339 & 0.372 \\
\multirow{-4}{*}{\cellcolor[HTML]{F1E5FF}\rotatebox{90}{ETTm2}} & 720 & 0.673 & 0.570 & \cellcolor[HTML]{6D9EEB}1.650 & \cellcolor[HTML]{A2C2F3}1.414 & {\color[HTML]{FF0000} \textbf{0.408}} & {\color[HTML]{FF0000} \textbf{0.403}} & {\color[HTML]{0000FF} \underline{ 0.414}} & {\color[HTML]{0000FF} \underline{ 0.413}} & 0.675 & 0.587 & 0.554 & 0.522 & 0.421 & 0.415 & 0.417 & {\color[HTML]{0000FF} \underline{ 0.413}} & 0.433 & 0.432 \\
\hline
\cellcolor[HTML]{F1E5FF} & 96 & {\color[HTML]{FF0000} \textbf{0.317}} & {\color[HTML]{FF0000} \textbf{0.370}} & \cellcolor[HTML]{F4CCCC}0.932 & \cellcolor[HTML]{F4CCCC}0.989 & 0.340 & {\color[HTML]{0000FF} \underline{ 0.374}} & 0.340 & 0.391 & 0.397 & 0.437 & {\color[HTML]{0000FF} \underline{ 0.333}} & 0.387 & 0.358 & 0.397 & 0.476 & 0.458 & 0.346 & 0.388 \\
\cellcolor[HTML]{F1E5FF} & 192 & {\color[HTML]{0000FF} \underline{ 0.420}} & {\color[HTML]{0000FF} \underline{ 0.433}} & \cellcolor[HTML]{F5F9FE}1.045 & \cellcolor[HTML]{F5F9FE}1.046 & {\color[HTML]{FF0000} \textbf{0.402}} & {\color[HTML]{FF0000} \textbf{0.414}} & 0.430 & 0.439 & 0.520 & 0.504 & 0.477 & 0.476 & 0.429 & 0.439 & 0.512 & 0.493 & 0.456 & 0.452 \\
\cellcolor[HTML]{F1E5FF} & 336 & 0.536 & 0.507 & \cellcolor[HTML]{D6E4FA}1.186 & \cellcolor[HTML]{E4EDFC}1.122 & {\color[HTML]{FF0000} \textbf{0.452}} & {\color[HTML]{FF0000} \textbf{0.452}} & 0.485 & {\color[HTML]{0000FF} \underline{ 0.479}} & 0.626 & 0.559 & 0.594 & 0.541 & 0.496 & 0.487 & 0.552 & 0.551 & {\color[HTML]{0000FF} \underline{ 0.482}} & 0.486 \\
\multirow{-4}{*}{\cellcolor[HTML]{F1E5FF}\rotatebox{90}{ETTh2}} & 720 & 0.719 & 0.601 & \cellcolor[HTML]{82ACEE}1.556 & \cellcolor[HTML]{C0D5F7}1.284 & {\color[HTML]{FF0000} \textbf{0.462}} & {\color[HTML]{FF0000} \textbf{0.468}} & 0.500 & 0.497 & 0.863 & 0.672 & 0.831 & 0.657 & {\color[HTML]{0000FF} \underline{ 0.463}} & {\color[HTML]{0000FF} \underline{ 0.474}} & 0.562 & 0.560 & 0.515 & 0.511 \\
\hline
\cellcolor[HTML]{F1E5FF} & 96 & 0.888 & 0.518 & \cellcolor[HTML]{90B5F0}1.497 & \cellcolor[HTML]{76A4ED}1.614 & {\color[HTML]{0000FF} \underline{ 0.593}} & {\color[HTML]{FF0000} \textbf{0.321}} & 0.607 & 0.392 & 0.615 & 0.391 & 0.650 & 0.396 & {\color[HTML]{FF0000} \textbf{0.587}} & 0.366 & 0.612 & {\color[HTML]{0000FF} \underline{ 0.338}} & 0.613 & 0.388 \\
\cellcolor[HTML]{F1E5FF} & 192 & 0.781 & 0.477 & \cellcolor[HTML]{C4D8F7}1.266 & \cellcolor[HTML]{A1C1F3}1.420 & 0.617 & {\color[HTML]{FF0000} \textbf{0.336}} & 0.621 & 0.399 & {\color[HTML]{0000FF} \underline{ 0.601}} & 0.382 & {\color[HTML]{FF0000} \textbf{0.598}} & 0.370 & 0.604 & 0.373 & 0.613 & {\color[HTML]{0000FF} \underline{ 0.340}} & 0.616 & 0.382 \\
\cellcolor[HTML]{F1E5FF} & 336 & 0.786 & 0.484 & \cellcolor[HTML]{C7DAF8}1.250 & \cellcolor[HTML]{9CBEF2}1.440 & 0.629 & {\color[HTML]{0000FF} \underline{ 0.336}} & 0.622 & 0.396 & {\color[HTML]{0000FF} \underline{ 0.613}} & 0.386 & {\color[HTML]{FF0000} \textbf{0.605}} & 0.373 & 0.621 & 0.383 & 0.618 & {\color[HTML]{FF0000} \textbf{0.328}} & 0.622 & 0.337 \\
\multirow{-4}{*}{\cellcolor[HTML]{F1E5FF}\rotatebox{90}{Traffic}} & 720 & 0.813 & 0.497 & \cellcolor[HTML]{C3D7F7}1.270 & \cellcolor[HTML]{A1C1F3}1.420 & 0.640 & {\color[HTML]{FF0000} \textbf{0.350}} & {\color[HTML]{0000FF} \underline{ 0.632}} & 0.396 & 0.658 & 0.407 & 0.645 & 0.394 & {\color[HTML]{FF0000} \textbf{0.626}} & 0.382 & 0.653 & {\color[HTML]{0000FF} \underline{ 0.355}} & 0.660 & 0.408 \\
\hline
\cellcolor[HTML]{F1E5FF} & 96 & 0.222 & 0.276 & \cellcolor[HTML]{BED4F7}1.291 & \cellcolor[HTML]{C6D9F8}1.255 & {\color[HTML]{FF0000} \textbf{0.172}} & {\color[HTML]{FF0000} \textbf{0.220}} & 0.197 & 0.281 & 0.182 & 0.242 & 0.196 & 0.255 & 0.217 & 0.296 & {\color[HTML]{0000FF} \underline{ 0.173}} & {\color[HTML]{0000FF} \underline{ 0.223}} & 0.266 & 0.336 \\
\cellcolor[HTML]{F1E5FF} & 192 & 0.265 & 0.311 & \cellcolor[HTML]{D0E0F9}1.210 & \cellcolor[HTML]{D4E3FA}1.192 & {\color[HTML]{FF0000} \textbf{0.219}} & {\color[HTML]{FF0000} \textbf{0.261}} & 0.237 & 0.312 & {\color[HTML]{0000FF} \underline{ 0.227}} & 0.287 & 0.237 & 0.296 & 0.276 & 0.336 & 0.245 & {\color[HTML]{0000FF} \underline{ 0.285}} & 0.307 & 0.367 \\
\cellcolor[HTML]{F1E5FF} & 336 & 0.302 & 0.345 & \cellcolor[HTML]{EEF4FD}1.079 & \cellcolor[HTML]{E3ECFC}1.127 & {\color[HTML]{FF0000} \textbf{0.280}} & {\color[HTML]{FF0000} \textbf{0.306}} & 0.298 & 0.353 & {\color[HTML]{0000FF} \underline{ 0.282}} & {\color[HTML]{0000FF} \underline{ 0.334}} & 0.283 & 0.335 & 0.339 & 0.380 & 0.321 & 0.338 & 0.359 & 0.395 \\
\multirow{-4}{*}{\cellcolor[HTML]{F1E5FF}\rotatebox{90}{Weather}} & 720 & 0.375 & 0.406 & \cellcolor[HTML]{F9FBFF}1.027 & \cellcolor[HTML]{E2ECFB}1.131 & 0.365 & {\color[HTML]{0000FF} \underline{ 0.359}} & {\color[HTML]{0000FF} \underline{ 0.352}} & {\color[HTML]{FF0000} \textbf{0.288}} & {\color[HTML]{0000FF} \underline{ 0.352}} & 0.386 & {\color[HTML]{FF0000} \textbf{0.345}} & 0.381 & 0.403 & 0.428 & 0.414 & 0.410 & 0.419 & 0.428\\
\hline
\end{tabular}
\end{adjustbox}
\vspace{0.15cm} ~

\footnotesize{(*)
The \textcolor{red}{ \bf red bold} text indicates the best amongst all methods in Table \ref{tab:finetune}, and the \textcolor{blue}{ \underline{blue underlined}} text indicate the second best method. The datasets within the pretrain collection are highlighted with yellow color while the others highlighted purple. In the Ratio column, the numbers highlighted red indicates the settings where the pretrained model has lower metrics than TimesNet. The blue color scale indicates the settings where the pretrained test metrics are worse than TimesNet test results. }
\end{table*}

\subsection{Pretrain and Results} 
We choose a collection of pretraining datasets including ETTh1, ETTm1, Electricity and Exchange-Rate. The remaining datasets (ETTh2, ETTm2, Weather, Traffic, ILI) are reserved for fine-tuning and/or further testing the generalization ability of our model. Note that we do not compare the test performance for ILI data with the previous results because their setting of input and prediction lengths is different from other datasets. In order to handle the discrepancy within the sizes of the datasets, we design a pretraining collection such that it has equal number of data samples from each pretrain dataset. 
Note that we only choose a subset of features from Electricity data because it has much more than features other datasets.
The detailed description of this collection is in the Appendix.

In our experiment, we choose a simple linear layer for the encoder and the decoder of the model. We choose this simple architecture to better analyze the pretrain-finetune process and also because simple models have shown impressive performance in time series forecasting \citep{linearmodels, tran2023endtoend}. 
We train our model using PyTorch \citep{pytorch} with ADAM optimizer \citep{adam} for 10 training epochs. We choose the regularization parameter $\lambda = 0.1$ and the temperature parameter $\tau = 0.1$, the pretrain batch size is 512. The experiments are repeated 3 times. All the experiment details are in the Appendix. 
We compare the test errors (MSE and MAE) of our model with the following supervised learning models for time series forecasting: TimesNet \citep{timesnet}, ETSformer \citep{etsformer}, LightTS \citep{lightts}, DLinear \citep{linearmodels}, FEDformer \citep{fedformer}, Stationary \citep{stationary},
Autoformer \citep{autoformer}. 
Table \ref{tab:pretrain} shows a record of their test results. 

Note that we do not remove the last layer of our model before finetuning and the pretrained model can be tested on all of the datasets. Our pretrain results is reported in Table \ref{tab:pretrain}. In this table, we also report the ratios of the test errors between our method and TimesNet, a state-of-the-art model for time series processing.

\textbf{Results.} Table \ref{tab:pretrain} shows that our pretrained model has promising generalization results in most of the pretraining datasets. For ETTh1, the model perform only slightly worse than TimesNet i.e. the difference is $3.4 \%$ in average compared to TimesNet accuracy. The pretrain model has very good generalization in Exchange-Rate dataset, which is better than TimesNet in 7/8 settings and even better than all the supervised methods in the long term predictions (336 and 720 forward time steps). We argue that the pretrain process acts as a regularization for the Exchange-Rate dataset in this case. On the other hand, for ETTm1 and Electricity datasets the pretrained model is worse than TimesNet (the difference in average is $19.4\%$ and $30.7\%$, respectively). However, that is not surprising because the supervised models are trained only specifically on that dataset. 

For the other datasets which the pretrained models did not have access, it is reasonable that the test performance is worse than the datasets within the pretrain collection. The Weather and ETTh2 datasets have the best generalization with an average of $16.4\%$ difference and  $14.5\%$ difference compared to TimesNet method.

\subsection{Similarity Results}
In this section, we show the estimated probability from the pretrained model. We compute this metric using our collection of pretrain samples, and averaging over the finetune samples. We report the percentages in Table \ref{tab:prob}.

\begin{table}[h]
\caption{The chance that a finetune dataset is similar to one of the pretrain datasets, estimated by our pretrained model. The color scale highlights the high chances in red and low changes in blue. \\
}\label{tab:prob}
\centering
\begin{tabular}{|c|m{2cm}m{2cm}m{2cm}m{2cm}|}
\hline
 Finetune & \multicolumn{4}{c|}{\% of similarity to the pretrain datasets} \\
datasets  & \cellcolor[HTML]{FFF7E7}ETTh1 & \cellcolor[HTML]{FFF7E7}ETTm1 & \cellcolor[HTML]{FFF7E7}Exchange & \cellcolor[HTML]{FFF7E7}Electricity \\
\hline
\cellcolor[HTML]{FFF7E7}ETTh1 & \cellcolor[HTML]{EDA4A4}67.62 & \cellcolor[HTML]{C3D7F6}11.86 & \cellcolor[HTML]{C3D7F6}11.82 & \cellcolor[HTML]{ACC8F3}8.70 \\
\cellcolor[HTML]{FFF7E7}ETTm1 & \cellcolor[HTML]{C1D6F6}11.55 & \cellcolor[HTML]{F2BFBF}53.95 & \cellcolor[HTML]{FDF1F1}27.58 & \cellcolor[HTML]{9FBFF1}6.93 \\
\cellcolor[HTML]{FFF7E7}Exchange & \cellcolor[HTML]{78A5EC}1.56 & \cellcolor[HTML]{B6CEF5}10.01 & \cellcolor[HTML]{E57E7E}87.79 & \cellcolor[HTML]{71A1EB}0.65 \\
\cellcolor[HTML]{FFF7E7}~~~~~~Electricity~~~~~~ & \cellcolor[HTML]{DBE7FA}15.20 & \cellcolor[HTML]{FFFCFC}22.08 & \cellcolor[HTML]{F2BFBF}53.75 & \cellcolor[HTML]{AEC9F3}8.97 \\
\cellcolor[HTML]{F1E5FF}ETTh2 & \cellcolor[HTML]{FBEBEB}30.87 & \cellcolor[HTML]{FEFAFA}22.65 & \cellcolor[HTML]{FBEBEB}30.49 & \cellcolor[HTML]{E1EBFB}16.00 \\
\cellcolor[HTML]{F1E5FF}ETTm2 & \cellcolor[HTML]{DFEAFA}15.71 & \cellcolor[HTML]{FCECEC}30.10 & \cellcolor[HTML]{F6D2D2}44.00 & \cellcolor[HTML]{B7CFF5}10.19 \\
\cellcolor[HTML]{F1E5FF}Traffic & \cellcolor[HTML]{F7D7D7}41.40 & \cellcolor[HTML]{FFFCFC}21.66 & \cellcolor[HTML]{EBF2FC}17.39 & \cellcolor[HTML]{FBFCFE}19.54 \\
\cellcolor[HTML]{F1E5FF}Weather & \cellcolor[HTML]{70A0EB}0.52 & \cellcolor[HTML]{A3C1F2}7.4 & \cellcolor[HTML]{E47676}91.87 & \cellcolor[HTML]{6E9FEB}0.22 \\
\cellcolor[HTML]{F1E5FF}ILI & \cellcolor[HTML]{FDF4F4}25.97 & \cellcolor[HTML]{FAE5E5}34.07 & \cellcolor[HTML]{FBEBEB}30.93 & \cellcolor[HTML]{AEC9F4}9.04\\
\hline
\end{tabular}
\end{table}
This analysis shows that the pretrain model predicts well for the three datasets ETTh1, ETTm1 and Exchange. However, the model cannot classify Electricity data and this fact explains the bad generalization error in Table \ref{tab:pretrain}. The Exchange-Rate data is predicted with a high probability and the model shows good generalization. Among the other finetune datasets, a related phenomenon appears in Weather data: high probability of similarity and good generalization. ETTh2 datasets has good metrics since it is close to ETTh1 dataset, while ETTm2 has higher probability in Exchange data class. This confirms our finding that the model prediction and contrastive representation learning complement each other. Another observation from this experiment is that the correct classifications do not varies much between different sample batches.

\begin{table*}[ht]
\caption{Comparisons of the test performance from our finetuned model and other supervised learning models**.\\}\label{tab:finetune}
\begin{adjustbox}{width=0.99\textwidth}
\begin{tabular}{|c|c|cc|cc|cc|cc|cc|cc|cc|cc|cc|}
\hline
 & Models & \multicolumn{2}{c|}{Our FT model} & \multicolumn{2}{c|}{Ratio} & \multicolumn{2}{c|}{TimesNet} & \multicolumn{2}{c|}{ETSformer} & \multicolumn{2}{c|}{LightTS} & \multicolumn{2}{c|}{DLinear} & \multicolumn{2}{c|}{FEDformer} & \multicolumn{2}{c|}{Stationary} & \multicolumn{2}{c|}{Autoformer} \\

\multirow{-2}{*}{{Data}} & Metric & MSE & MAE & MSE & MAE & MSE & MAE & MSE & MAE & MSE & MAE & MSE & MAE & MSE & MAE & MSE & MAE & MSE & MAE \\
\hline
\cellcolor[HTML]{FFF7E7} & 96 & 0.347 & {\color[HTML]{0000FF} \underline{0.373}} & \cellcolor[HTML]{F8FBFE}1.027 & \cellcolor[HTML]{F4CCCC}0.995 & {\color[HTML]{FF0000} \textbf{0.338}} & 0.375 & 0.375 & 0.398 & 0.374 & 0.400 & {\color[HTML]{0000FF} \underline{0.345}} & {\color[HTML]{FF0000} \textbf{0.372}} & 0.379 & 0.419 & 0.386 & 0.398 & 0.505 & 0.475 \\
\cellcolor[HTML]{FFF7E7} & 192 & 0.384 & 0.393 & \cellcolor[HTML]{F8FBFE}1.027 & \cellcolor[HTML]{FBFDFF}1.016 & {\color[HTML]{FF0000} \textbf{0.374}} & {\color[HTML]{FF0000} \textbf{0.387}} & 0.408 & 0.410 & 0.400 & 0.407 & {\color[HTML]{0000FF} \underline{0.380}} & {\color[HTML]{0000FF} \underline{0.389}} & 0.426 & 0.441 & 0.459 & 0.444 & 0.553 & 0.496 \\
\cellcolor[HTML]{FFF7E7} & 336 & 0.414 & 0.415 & \cellcolor[HTML]{FDFEFF}1.010 & \cellcolor[HTML]{FDFEFF}1.010 & {\color[HTML]{FF0000} \textbf{0.410}} & {\color[HTML]{FF0000} \textbf{0.411}} & 0.435 & 0.428 & 0.438 & 0.438 & {\color[HTML]{0000FF} \underline{0.413}} & {\color[HTML]{0000FF} \underline{0.413}} & 0.445 & 0.459 & 0.495 & 0.464 & 0.621 & 0.537 \\
\multirow{-4}{*}{\cellcolor[HTML]{FFF7E7}\rotatebox{90}{ETTm1}} & 720 & {\color[HTML]{FF0000} \textbf{0.473}} & {\color[HTML]{0000FF} \underline{0.451}} & \cellcolor[HTML]{F4CCCC}0.990 & \cellcolor[HTML]{FFFFFF}1.002 & 0.478 & {\color[HTML]{FF0000} \textbf{0.450}} & 0.499 & 0.462 & 0.527 & 0.502 & {\color[HTML]{0000FF} \underline{0.474}} & 0.453 & 0.543 & 0.490 & 0.585 & 0.516 & 0.671 & 0.561 \\
\hline
\cellcolor[HTML]{FFF7E7} & 96 & 0.385 & {\color[HTML]{FF0000} \textbf{0.398}} & \cellcolor[HTML]{FFFFFF}1.003 & \cellcolor[HTML]{F4CCCC}0.990 & {\color[HTML]{0000FF} \underline{0.384}} & 0.402 & 0.494 & 0.479 & 0.424 & 0.432 & 0.386 & {\color[HTML]{0000FF} \underline{0.400}} & {\color[HTML]{FF0000} \textbf{0.376}} & 0.419 & 0.513 & 0.491 & 0.449 & 0.459 \\
\cellcolor[HTML]{FFF7E7} & 192 & {\color[HTML]{0000FF} \underline{0.432}} & {\color[HTML]{FF0000} \textbf{0.425}} & \cellcolor[HTML]{F4CCCC}0.991 & \cellcolor[HTML]{F4CCCC}0.991 & 0.436 & {\color[HTML]{0000FF} \underline{0.429}} & 0.538 & 0.504 & 0.475 & 0.462 & 0.437 & 0.432 & {\color[HTML]{FF0000} \textbf{0.420}} & 0.448 & 0.534 & 0.504 & 0.500 & 0.482 \\
\cellcolor[HTML]{FFF7E7} & 336 & {\color[HTML]{0000FF} \underline{0.473}} & {\color[HTML]{FF0000} \textbf{0.448}} & \cellcolor[HTML]{F4CCCC}0.963 & \cellcolor[HTML]{F4CCCC}0.955 & 0.491 & 0.469 & 0.574 & 0.521 & 0.518 & 0.488 & 0.481 & {\color[HTML]{0000FF} \underline{0.459}} & {\color[HTML]{FF0000} \textbf{0.459}} & 0.465 & 0.588 & 0.535 & 0.521 & 0.496 \\
\multirow{-4}{*}{\cellcolor[HTML]{FFF7E7}\rotatebox{90}{ETTh1}} & 720 & {\color[HTML]{FF0000} \textbf{0.492}} & {\color[HTML]{FF0000} \textbf{0.490}} & \cellcolor[HTML]{F4CCCC}0.944 & \cellcolor[HTML]{F4CCCC}0.980 & 0.521 & {\color[HTML]{0000FF} \underline{0.500}} & 0.562 & 0.535 & 0.547 & 0.533 & 0.519 & 0.516 & {\color[HTML]{0000FF} \underline{0.506}} & 0.507 & 0.643 & 0.616 & 0.514 & 0.512 \\
\hline
\cellcolor[HTML]{FFF7E7} & 96 & {\color[HTML]{FF0000} \textbf{0.081}} & {\color[HTML]{FF0000} \textbf{0.204}} & \cellcolor[HTML]{F4CCCC}0.757 & \cellcolor[HTML]{F4CCCC}0.872 & 0.107 & 0.234 & {\color[HTML]{0000FF} \underline{0.085}} & {\color[HTML]{FF0000} \textbf{0.204}} & 0.116 & 0.262 & 0.088 & 0.218 & 0.148 & 0.278 & 0.111 & 0.237 & 0.197 & 0.323 \\
\cellcolor[HTML]{FFF7E7} & 192 & {\color[HTML]{FF0000} \textbf{0.164}} & {\color[HTML]{FF0000} \textbf{0.300}} & \cellcolor[HTML]{F4CCCC}0.726 & \cellcolor[HTML]{F4CCCC}0.872 & 0.226 & 0.344 & 0.182 & {\color[HTML]{0000FF} \underline{0.303}} & 0.215 & 0.359 & {\color[HTML]{0000FF} \underline{0.176}} & 0.315 & 0.271 & 0.380 & 0.219 & 0.335 & 0.300 & 0.369 \\
\cellcolor[HTML]{FFF7E7} & 336 & {\color[HTML]{FF0000} \textbf{0.295}} & {\color[HTML]{FF0000} \textbf{0.407}} & \cellcolor[HTML]{F4CCCC}0.804 & \cellcolor[HTML]{F4CCCC}0.908 & 0.367 & 0.448 & 0.348 & 0.428 & 0.377 & 0.466 & {\color[HTML]{0000FF} \underline{0.313}} & {\color[HTML]{0000FF} \underline{0.427}} & 0.460 & 0.500 & 0.421 & 0.476 & 0.509 & 0.524 \\
\multirow{-4}{*}{\cellcolor[HTML]{FFF7E7}\rotatebox{90}{Exchange}} & 720 & {\color[HTML]{FF0000} \textbf{0.535}} & {\color[HTML]{FF0000} \textbf{0.587}} & \cellcolor[HTML]{F4CCCC}0.555 & \cellcolor[HTML]{F4CCCC}0.787 & 0.964 & 0.746 & 1.025 & 0.774 & {\color[HTML]{0000FF} \underline{0.831}} & 0.699 & 0.839 & {\color[HTML]{0000FF} \underline{0.695}} & 1.195 & 0.841 & 1.092 & 0.769 & 1.447 & 0.941 \\
\hline
\cellcolor[HTML]{FFF7E7} & 96 & 0.197 & 0.282 & \cellcolor[HTML]{D0E0F9}1.173 & \cellcolor[HTML]{F5F9FE}1.037 & {\color[HTML]{FF0000} \textbf{0.168}} & {\color[HTML]{FF0000} \textbf{0.272}} & 0.187 & 0.304 & 0.207 & 0.307 & 0.197 & 0.282 & 0.193 & 0.308 & {\color[HTML]{0000FF} \underline{0.169}} & {\color[HTML]{0000FF} \underline{0.273}} & 0.201 & 0.317 \\
\cellcolor[HTML]{FFF7E7} & 192 & 0.195 & {\color[HTML]{FF0000} \textbf{0.282}} & \cellcolor[HTML]{EFF5FD}1.060 & \cellcolor[HTML]{F4CCCC}0.976 & {\color[HTML]{0000FF} \underline{0.184}} & 0.289 & 0.199 & 0.315 & 0.213 & 0.316 & 0.196 & {\color[HTML]{0000FF} \underline{0.285}} & 0.201 & 0.315 & {\color[HTML]{FF0000} \textbf{0.182}} & 0.286 & 0.222 & 0.334 \\
\cellcolor[HTML]{FFF7E7} & 336 & 0.207 & {\color[HTML]{FF0000} \textbf{0.296}} & \cellcolor[HTML]{F3F7FE}1.045 & \cellcolor[HTML]{F4CCCC}0.987 & {\color[HTML]{FF0000} \textbf{0.198}} & {\color[HTML]{0000FF} \underline{0.300}} & 0.212 & 0.329 & 0.230 & 0.333 & 0.209 & 0.301 & 0.214 & 0.329 & {\color[HTML]{0000FF} \underline{0.200}} & 0.304 & 0.231 & 0.338 \\
\multirow{-4}{*}{\cellcolor[HTML]{FFF7E7}\rotatebox{90}{Electricity}} & 720 & 0.242 & {\color[HTML]{FF0000} \textbf{0.229}} & \cellcolor[HTML]{E4EDFC}1.100 & \cellcolor[HTML]{F4CCCC}0.716 & {\color[HTML]{FF0000} \textbf{0.220}} & {\color[HTML]{0000FF} \underline{0.320}} & 0.233 & 0.345 & 0.265 & 0.360 & 0.245 & 0.333 & 0.246 & 0.355 & {\color[HTML]{0000FF} \underline{0.222}} & 0.321 & 0.254 & 0.361 \\
\hline
\cellcolor[HTML]{F1E5FF} & 96 & 0.196 & 0.294 & \cellcolor[HTML]{F2F7FE}1.048 & \cellcolor[HTML]{E4EDFC}1.101 & {\color[HTML]{FF0000} \textbf{0.187}} & {\color[HTML]{FF0000} \textbf{0.267}} & {\color[HTML]{0000FF} \underline{0.189}} & 0.280 & 0.209 & 0.308 & 0.193 & 0.292 & 0.203 & 0.287 & 0.192 & {\color[HTML]{0000FF} \underline{0.274}} & 0.255 & 0.339 \\
\cellcolor[HTML]{F1E5FF} & 192 & 0.266 & 0.342 & \cellcolor[HTML]{EDF3FD}1.068 & \cellcolor[HTML]{E2ECFB}1.107 & {\color[HTML]{FF0000} \textbf{0.249}} & {\color[HTML]{FF0000} \textbf{0.309}} & {\color[HTML]{0000FF} \underline{0.253}} & {\color[HTML]{0000FF} \underline{0.319}} & 0.311 & 0.382 & 0.284 & 0.362 & 0.269 & 0.328 & 0.280 & 0.339 & 0.281 & 0.340 \\
\cellcolor[HTML]{F1E5FF} & 336 & 0.365 & 0.412 & \cellcolor[HTML]{DAE7FA}1.137 & \cellcolor[HTML]{D0E0F9}1.174 & {\color[HTML]{0000FF} \underline{0.321}} & {\color[HTML]{FF0000} \textbf{0.351}} & {\color[HTML]{FF0000} \textbf{0.314}} & {\color[HTML]{0000FF} \underline{0.357}} & 0.442 & 0.466 & 0.369 & 0.427 & 0.325 & 0.366 & 0.334 & 0.361 & 0.339 & 0.372 \\
\multirow{-4}{*}{\cellcolor[HTML]{F1E5FF}\rotatebox{90}{ETTm2}} & 720 & 0.492 & 0.485 & \cellcolor[HTML]{C7DAF8}1.206 & \cellcolor[HTML]{C8DAF8}1.203 & {\color[HTML]{FF0000} \textbf{0.408}} & {\color[HTML]{FF0000} \textbf{0.403}} & {\color[HTML]{0000FF} \underline{0.414}} & {\color[HTML]{0000FF} \underline{0.413}} & 0.675 & 0.587 & 0.554 & 0.522 & 0.421 & 0.415 & 0.417 & {\color[HTML]{0000FF} \underline{0.413}} & 0.433 & 0.432 \\
\hline
\cellcolor[HTML]{F1E5FF} & 96 & {\color[HTML]{FF0000} \textbf{0.316}} & {\color[HTML]{FF0000} \textbf{0.372}} & \cellcolor[HTML]{F4CCCC}0.929 & \cellcolor[HTML]{F4CCCC}0.995 & 0.340 & {\color[HTML]{0000FF} \underline{0.374}} & 0.340 & 0.391 & 0.397 & 0.437 & {\color[HTML]{0000FF} \underline{0.333}} & 0.387 & 0.358 & 0.397 & 0.476 & 0.458 & 0.346 & 0.388 \\
\cellcolor[HTML]{F1E5FF} & 192 & {\color[HTML]{0000FF} \underline{0.419}} & {\color[HTML]{0000FF} \underline{0.433}} & \cellcolor[HTML]{F4F8FE}1.042 & \cellcolor[HTML]{F3F7FE}1.046 & {\color[HTML]{FF0000} \textbf{0.402}} & {\color[HTML]{FF0000} \textbf{0.414}} & 0.430 & 0.439 & 0.520 & 0.504 & 0.477 & 0.476 & 0.429 & 0.439 & 0.512 & 0.493 & 0.456 & 0.452 \\
\cellcolor[HTML]{F1E5FF} & 336 & 0.536 & 0.507 & \cellcolor[HTML]{CDDEF9}1.186 & \cellcolor[HTML]{DEE9FB}1.122 & {\color[HTML]{FF0000} \textbf{0.452}} & {\color[HTML]{FF0000} \textbf{0.452}} & 0.485 & {\color[HTML]{0000FF} \underline{0.479}} & 0.626 & 0.559 & 0.594 & 0.541 & 0.496 & 0.487 & 0.552 & 0.551 & {\color[HTML]{0000FF} \underline{0.482}} & 0.486 \\
\multirow{-4}{*}{\cellcolor[HTML]{F1E5FF}\rotatebox{90}{ETTh2}} & 720 & 0.708 & 0.543 & \cellcolor[HTML]{6D9EEB}1.532 & \cellcolor[HTML]{D4E2F9}1.160 & {\color[HTML]{FF0000} \textbf{0.462}} & {\color[HTML]{FF0000} \textbf{0.468}} & 0.500 & 0.497 & 0.863 & 0.672 & 0.831 & 0.657 & {\color[HTML]{0000FF} \underline{0.463}} & {\color[HTML]{0000FF} \underline{0.474}} & 0.562 & 0.560 & 0.515 & 0.511 \\
\hline
\cellcolor[HTML]{F1E5FF} & 96 & 0.664 & 0.410 & \cellcolor[HTML]{DFEAFB}1.120 & \cellcolor[HTML]{B3CDF5}1.277 & {\color[HTML]{0000FF} \underline{0.593}} & {\color[HTML]{FF0000} \textbf{0.321}} & 0.607 & 0.392 & 0.615 & 0.391 & 0.650 & 0.396 & {\color[HTML]{FF0000} \textbf{0.587}} & 0.366 & 0.612 & {\color[HTML]{0000FF} \underline{0.338}} & 0.613 & 0.388 \\
\cellcolor[HTML]{F1E5FF} & 192 & 0.606 & 0.380 & \cellcolor[HTML]{F4CCCC}0.982 & \cellcolor[HTML]{DCE8FB}1.131 & 0.617 & {\color[HTML]{FF0000} \textbf{0.336}} & 0.621 & 0.399 & {\color[HTML]{0000FF} \underline{0.601}} & 0.382 & {\color[HTML]{FF0000} \textbf{0.598}} & 0.370 & 0.604 & 0.373 & 0.613 & {\color[HTML]{0000FF} \underline{0.340}} & 0.616 & 0.382 \\
\cellcolor[HTML]{F1E5FF} & 336 & {\color[HTML]{0000FF} \underline{0.608}} & 0.378 & \cellcolor[HTML]{F4CCCC}0.967 & \cellcolor[HTML]{DDE9FB}1.125 & 0.629 & {\color[HTML]{0000FF} \underline{0.336}} & 0.622 & 0.396 & 0.613 & 0.386 & {\color[HTML]{FF0000} \textbf{0.605}} & 0.373 & 0.621 & 0.383 & 0.618 & {\color[HTML]{FF0000} \textbf{0.328}} & 0.622 & 0.337 \\
\multirow{-4}{*}{\cellcolor[HTML]{F1E5FF}\rotatebox{90}{Traffic}} & 720 & 0.647 & 0.398 & \cellcolor[HTML]{FDFEFF}1.011 & \cellcolor[HTML]{DAE7FA}1.137 & 0.640 & {\color[HTML]{FF0000} \textbf{0.350}} & {\color[HTML]{0000FF} \underline{0.632}} & 0.396 & 0.658 & 0.407 & 0.645 & 0.394 & {\color[HTML]{FF0000} \textbf{0.626}} & 0.382 & 0.653 & {\color[HTML]{0000FF} \underline{0.355}} & 0.660 & 0.408 \\
\hline
\cellcolor[HTML]{F1E5FF} & 96 & 0.194 & 0.251 & \cellcolor[HTML]{DCE8FB}1.128 & \cellcolor[HTML]{D9E6FA}1.141 & {\color[HTML]{FF0000} \textbf{0.172}} & {\color[HTML]{FF0000} \textbf{0.220}} & 0.197 & 0.281 & 0.182 & 0.242 & 0.196 & 0.255 & 0.217 & 0.296 & {\color[HTML]{0000FF} \underline{0.173}} & {\color[HTML]{0000FF} \underline{0.223}} & 0.266 & 0.336 \\
\cellcolor[HTML]{F1E5FF} & 192 & 0.235 & 0.291 & \cellcolor[HTML]{EBF2FD}1.073 & \cellcolor[HTML]{E0EBFB}1.115 & {\color[HTML]{FF0000} \textbf{0.219}} & {\color[HTML]{FF0000} \textbf{0.261}} & 0.237 & 0.312 & {\color[HTML]{0000FF} \underline{0.227}} & 0.287 & 0.237 & 0.296 & 0.276 & 0.336 & 0.245 & {\color[HTML]{0000FF} \underline{0.285}} & 0.307 & 0.367 \\
\cellcolor[HTML]{F1E5FF} & 336 & {\color[HTML]{0000FF} \underline{0.281}} & {\color[HTML]{0000FF} \underline{0.327}} & \cellcolor[HTML]{FFFFFF}1.004 & \cellcolor[HTML]{EDF3FD}1.069 & {\color[HTML]{FF0000} \textbf{0.280}} & {\color[HTML]{FF0000} \textbf{0.306}} & 0.298 & 0.353 & 0.282 & 0.334 & 0.283 & 0.335 & 0.339 & 0.380 & 0.321 & 0.338 & 0.359 & 0.395 \\
\multirow{-4}{*}{\cellcolor[HTML]{F1E5FF}\rotatebox{90}{Weather}} & 720 & {\color[HTML]{FF0000} \textbf{0.344}} & 0.368 & \cellcolor[HTML]{F4CCCC}0.942 & \cellcolor[HTML]{F9FBFF}1.025 & 0.365 & {\color[HTML]{0000FF} \underline{0.359}} & 0.352 & {\color[HTML]{FF0000} \textbf{0.288}} & 0.352 & 0.386 & {\color[HTML]{0000FF} \underline{0.345}} & 0.381 & 0.403 & 0.428 & 0.414 & 0.410 & 0.419 & 0.428\\
\hline
\end{tabular}
\end{adjustbox}
\vspace{0.15cm} ~

\small{
(**) The \textcolor{red}{ \bf red bold} text indicates the best amongst all methods, and the \textcolor{blue}{ \underline{blue underlined}} text indicate the second best method. The datasets within the pretrain collection are highlighted with yellow color while the others highlighted purple. In the Ratio column, the numbers highlighted red indicate the settings where the finetuned model has lower metrics than TimesNet. The blue color scale indicates the settings where the finetuned test metrics are worse than TimesNet.}
\end{table*}

\subsection{Finetune and Results}
For the finetune step, we estimate the probabilities using a batch of pretrain samples (512 samples) and update the model weights using the finetune loss on that batch. In this stage, the pretrain batch of samples has equal proportions of each datasets. On each finetune datasets, we finetune the model using 50\% of the training set compared to supervised approach (split with validation and test sets in chronological order by the ratio of 7:1:2) for 10 epochs. In Table \ref{tab:finetune}, we report the test performance of the models which yields the best validation result and the corresponding ratio with TimesNet. Note that the test batch size is 32, consistent with prior practice \citep{autoformer}. 

\textbf{Results.} Table \ref{tab:finetune} shows that the finetuned model generalizes better than the supervised methods for two datasets ETTh1 and Exchange-Rate. Exchange-Rate data shows an impressive improvement which yields better results than supervised methods in all settings. This is consistent with our probabilistic result that the model predicts the Exchange-Rate and ETTh1 data the best amongst other datasets. In ETTm1 and Electricity datasets, the model is comparable with or only slightly worse than TimesNet where the difference in average is $0.9\%$ and $1.1\%$, respectively. 

When finetune with new datasets, the model shows competitive performance for Traffic and Weather datasets (with $6.2\%$ and $9.4\%$ worse in average difference). The ETTh2 and ETTm2 datasets follow with an average of $12.6\%$ and $13.0\%$ worse in difference. 
Given the fact that the model is trained with a different collection of datasets, this experiment shows promising results for our pretrain-finetune approach. 
\section{Further Analysis}

\subsection{Variations Analysis}

In this section, we analyze some variations of our model. In the first variation, instead of applying the contrastive loss directly on the representation $z$, we use another decoder to transform it to a representation $y$, then apply the contrastive learning on $y$. We also use a linear layer for the decoder transforming $z$ to $y$. For the second variation, we do not use a decoder for the prediction output (i.e. using an identity layer in Figure \ref{fig:des}(a) instead of the decoder). In the final variation, we replace the linear encoder and decoder by two-layer neural networks. 
We report the results in Table \ref{tab:var}. This analysis shows that the linear encoder-decoder is essential for the good generalization performance of our method. The first variation with two decoders is also able to capture the dynamic, its performance has $6.2\%$ difference in average compared to the implementation of contrastive learning directly on the representation $z$. The second variation shows that a decoder for the prediction output is needed. 
The full results are in the Appendix. 
\setlength{\tabcolsep}{5pt}
\begin{table}[ht]
\caption{Comparisons of the test performance from our finetuned model and the variations. We report the average MSE and MAE over four prediction lengths (96, 192, 336, 720) as in Table \ref{tab:finetune}.\\
}\label{tab:var}
\centering
\begin{tabular}{|c|rr|rr|rr|rr|}
\hline
 & \multicolumn{2}{c|}{Finetuned} & \multicolumn{2}{c|}{Variation 1} & \multicolumn{2}{c|}{\cellcolor[HTML]{FFFFFF}Variation 2} & \multicolumn{2}{c|}{Variation 3} \\

\multirow{-2}{*}{Data} & \multicolumn{1}{c}{MSE} & \multicolumn{1}{c|}{MAE} & \multicolumn{1}{c}{MSE} & \multicolumn{1}{c|}{MAE} & \multicolumn{1}{c}{MSE} & \multicolumn{1}{c|}{MAE} & \multicolumn{1}{c}{MSE} & \multicolumn{1}{c|}{MAE} \\
\hline
\cellcolor[HTML]{FFF7E7}ETTm1 & {\color[HTML]{FF0000} \textbf{0.405}} & {\color[HTML]{FF0000} \textbf{0.408}} & {\color[HTML]{0000FF} \underline{0.410}} & {\color[HTML]{0000FF} \underline{0.424}} & 0.473 & 0.461 & 0.480 & 0.465 \\
\cellcolor[HTML]{FFF7E7}ETTh1 & {\color[HTML]{FF0000} \textbf{0.446}} & {\color[HTML]{FF0000} \textbf{0.440}} & 0.482 & {\color[HTML]{0000FF} \underline{0.484}} & {\color[HTML]{0000FF} \underline{0.476}} & 0.528 & 0.518 & 0.525 \\
\cellcolor[HTML]{FFF7E7}Exchange & {\color[HTML]{FF0000} \textbf{0.269}} & {\color[HTML]{FF0000} \textbf{0.375}} & {\color[HTML]{FF0000} \textbf{0.269}} & {\color[HTML]{0000FF} \underline{0.396}} & {\color[HTML]{0000FF} \underline{0.281}} & 0.414 & 0.282 & 0.423 \\
\cellcolor[HTML]{FFF7E7}Electricity & {\color[HTML]{FF0000} \textbf{0.210}} & {\color[HTML]{FF0000} \textbf{0.272}} & 0.227 & {\color[HTML]{0000FF} \underline{0.298}} & {\color[HTML]{0000FF} \underline{0.221}} & 0.338 & 0.243 & 0.320 \\
\hline
\cellcolor[HTML]{F1E5FF}ETTm2 & {\color[HTML]{FF0000} \textbf{0.330}} & {\color[HTML]{FF0000} \textbf{0.383}} & {\color[HTML]{0000FF} \underline{0.356}} & {\color[HTML]{0000FF} \underline{0.418}} & 0.367 & {\color[HTML]{0000FF} \underline{0.418}} & 0.395 & 0.451 \\
\cellcolor[HTML]{F1E5FF}ETTh2 & {\color[HTML]{FF0000} \textbf{0.495}} & {\color[HTML]{FF0000} \textbf{0.464}} & {\color[HTML]{0000FF} \underline{0.514}} & {\color[HTML]{0000FF} \underline{0.498}} & 0.532 & 0.549 & 0.522 & 0.542 \\
\cellcolor[HTML]{F1E5FF}Traffic & {\color[HTML]{FF0000} \textbf{0.631}} & {\color[HTML]{FF0000} \textbf{0.392}} & {\color[HTML]{0000FF} \underline{0.643}} & {\color[HTML]{0000FF} \underline{0.412}} & 0.693 & 0.431 & 0.773 & 0.428 \\
\cellcolor[HTML]{F1E5FF}Weather & {\color[HTML]{FF0000} \textbf{0.264}} & {\color[HTML]{FF0000} \textbf{0.309}} & 0.282 & {\color[HTML]{0000FF} \underline{0.344}} & {\color[HTML]{0000FF} \underline{0.271}} & 0.368 & 0.310 & 0.354\\
\hline
\end{tabular}
\end{table}

\subsection{Parameters Analysis}
In this section, we analyze how the performance of the model changes when the model parameters $\lambda$ changes. A similar analysis for $\tau$ is delayed to the Appendix. Note that $\lambda$ and $\tau$ are the regularization factor and the temperature parameter of the model, respectively, which controls the supervised contrastive loss term. 

\begin{figure}[ht]
    \centering
    \includegraphics[width=0.85\textwidth]{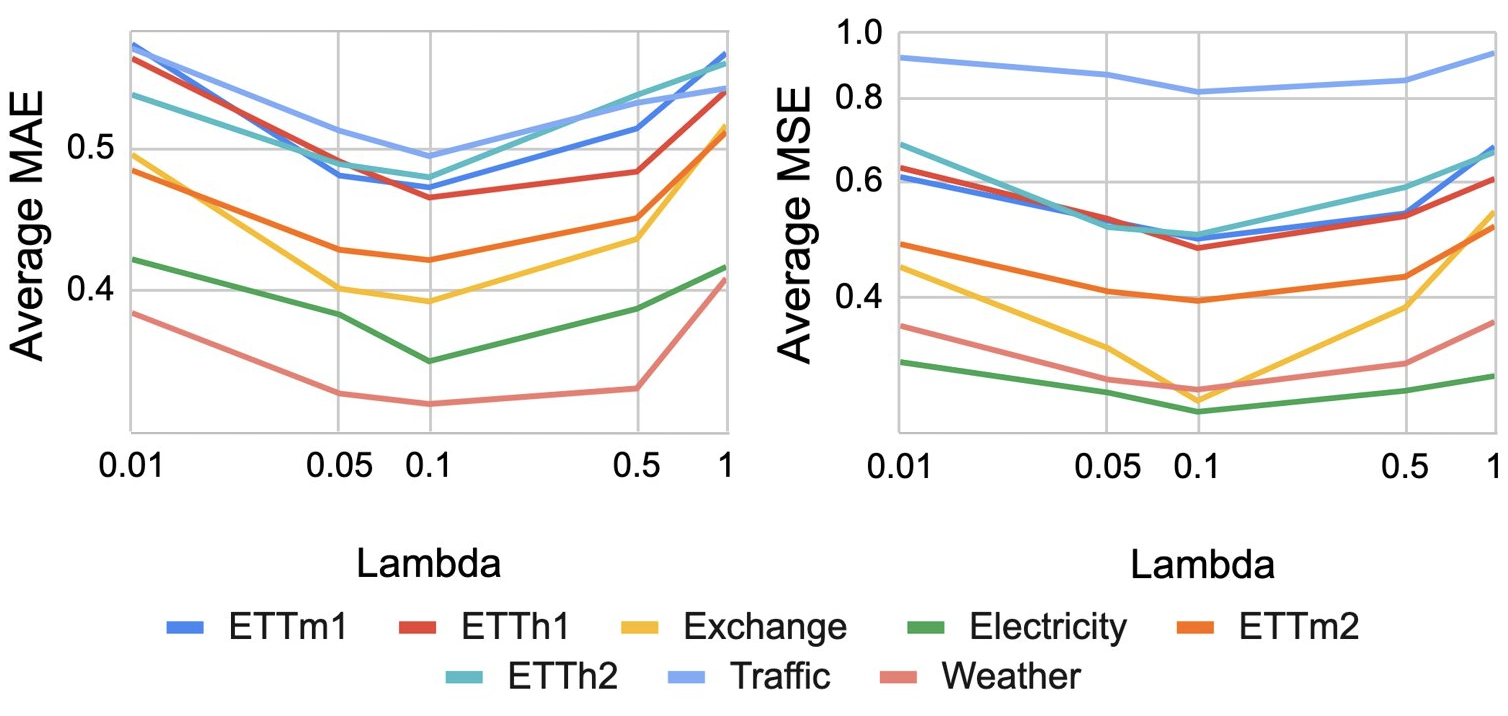}
    \caption{The average test performance of the pretrained model by parameter $\lambda$, for eights datasets. We plot both the $x$-axis and $y$-axis in logarithm scale to highlight the differences in accuracy. } 
    \label{fig:lamb}
\end{figure}

The choice of the regularization parameter $\lambda$ varies in $[0.01, 0.05, 0.1, 0.5, 1]$. We perform the test for eight datasets using the pretrained models and report the average errors (MAE and MSE) over the four prediction lengths in Figure \ref{fig:lamb}. We observe that in most datasets, the choice $\lambda = 0.1$ performs the best. 

\section{Conclusion}
In conclusion, our approach aims to address the discrepancy in pretrain-finetune time series and enhance the knowledge within foundation models training. We employ a pretraining procedure incorporating contrastive learning to distinguish features from different pretraining datasets. This supports the development of a probabilistic similarity metric, allows us to assess the likelihood of a univariate sample's similarity to one of the pretraining datasets.
We introduce a fine-tuning procedure designed to leverage the estimated probability. Our experiments demonstrate that our approach yields favorable results, with accuracy levels comparable to or in some cases outperform supervised models.

Future work in this direction offers promising problems. Addressing the inaccurate probability estimation is one of the interesting questions, which may requires further study into the dynamic of the pretraining datasets. There could be many potential reasons for this phenomenon: the two datasets may have similar dynamic that is difficult to distinguish, or the collective training with other datasets makes the model converges to some sub-optimal solutions. 
Another potential problem involves the discrepancy in a lower level: within each datasets. While we consider the simplified setting that features in the same datasets should be closer than the features from different datasets, it is still beneficial to take into account the potential dynamic variations within each dataset, and further apply that knowledge to improve the models.

\appendix
\section{Experiment Details}
\subsection{Pretrain Sample Collection}
We consider the time series forecasting problem where the model has the information of previous $I$ time steps and aims to predict the next $O$ future time steps. Thus every univariate sample has $I+O$ time steps. Note that we follow the standard experiment procedure as in \citep{autoformer}, using following real-world datasets: ETT, Traffic, Electricity, Weather, Exchange-Rate and ILI. The time series data is split into training, validation and test sets in chronological order by the ratio of 7:1:2 for all the data sets. To ensure fair comparison, our pretraining collection only contains the samples in the training proportions of the original datasets.  

We choose a collection of pretraining datasets including ETTh1, ETTm1, Electricity and Exchange-Rate. The remaining datasets (ETTh2, ETTm2, Weather, Traffic) are reserved for fine-tuning and/or further testing the generalization ability of our model. Note that in order to handle the discrepancy within the sizes of the datasets, we design a pretraining collection such that it has (approximately) the same number of data samples from each pretrain dataset.

The total number of (univariate) samples within each dataset scales proportionally with their number of features $d$ and total time steps $T$. ETTh1 and ETTm1 represents similar data, however their granularities are different: ETTh1 is recorded hourly while ETTm1 is collected every 15-minute interval. We choose the ETTm1 to be the base dataset, and we sample other datasets so that they have approximately the same number of samples as ETTm1. Since the total time steps of ETTh1 is 4 times less than ETTm1, we repeat the data ETTh1 for 4 times. 
Similarly, we repeat the data Exchange-Rate for 6 time since it has 8 features with a total time length of 7588 compared to the total length 69680 and 7 features of ETTm1. Finally, since the Electricity is too large, we choose a stride 2 to reduce the total length in half, and we only choose a subset of features (27 features) because Electricity data has much more than features other datasets (321). These 27 features have equally spaced indices of the original 321 features. Table \ref{tab:my_table2} describe our sampling procedure along with the descriptions of all the datasets in our experiments.


\setlength{\tabcolsep}{5pt}

\renewcommand{\arraystretch}{1.25}

\begin{table}[hb]
\centering
\caption{Description the datasets within the pretrain collection and other datasets used in our experiment.}
\label{tab:my_table2}
\begin{tabular}{|l|c|c|c|c|c|c|c|}
\hline
 &  &  &  & Used in &  &  & Number of features \\
\multirow{-2}{*}{Datasets} & \multirow{-2}{*}{$d$} & \multirow{-2}{*}{$T$} & \multirow{-2}{*}{Granularity} & pretrain data & \multirow{-2}{*}{Stride} & \multirow{-2}{*}{Repetition} & in pretrain \\
\hline
\rowcolor[HTML]{FFF7E7} 
ETTh1 & 7 & 17,420 & 1 hour & Yes & 1 & 4 & 7 \\
\rowcolor[HTML]{FFF7E7} 
ETTm1 & 7 & 69,680 & 15 min & Yes & 1 & 1 & 7 \\
\rowcolor[HTML]{FFF7E7} 
Electricity & 321 & 26,304 & 1 hour & Yes & 2 & 1 & 27 \\
\rowcolor[HTML]{FFF7E7} 
Exchange-Rate & 8 & 7,588 & 1 day & Yes & 1 & 6 & 8 \\
\hline
\rowcolor[HTML]{F1E5FF} 
ETTh2 & 7 & 17,420 & 1 hour & No &  &  &  \\
\rowcolor[HTML]{F1E5FF} 
ETTm2 & 7 & 69,680 & 15 min & No &  &  &  \\
\rowcolor[HTML]{F1E5FF} 
Traffic & 862 & 17,544 & 1 hour & No &  &  &  \\
\rowcolor[HTML]{F1E5FF} 
Weather & 21 & 52,696 & 10 min & No &  &  &  \\
\rowcolor[HTML]{F1E5FF} 
ILI & 7 & 966 & 1 week & No &  &  & \\
\hline
\end{tabular}
\end{table}

\subsection{Other experiment details}
In our experiment, we choose a simple linear layer (with bias) for the encoder and the decoder of the model. We choose this simple architecture to better analyze the pretrain-finetune process and also because linear models have shown impressive performance in time series forecasting \citep{linearmodels}. 
We test different dimensions for the representation of our model, for example we perform grid search with $\{48, 96, 192, 384\}$ for outputs 96 and 192, while we search in $\{180 ,360, 720, 1440\}$ for larger output of 720. We report the results where the representation space is half the output space, as it perform well in our experiment. Table \ref{tab:model} describes this choice and reports our model size. 

We pretrain our model using PyTorch \citep{pytorch} with ADAM optimizer \citep{adam} for 10 training epochs. We note that the size of our pretraining sample collection is approximately four times the size of ETTm1 (univariate) dataset. We choose the regularization parameter $\lambda = 0.1$ and the temperature parameter $\tau = 0.1$, the pretrain batch size is 512. The experiments are repeated 3 times.

\begin{table}[h]
\centering
\caption{Description of the model used in our pretrain and finetune experiments}
\label{tab:model}
\begin{tabular}{|c|c|c|c|}
\hline
Input  & Representation Dimension & Output & Model size  \\
\hline
96 & 48  & 96  & 9360 \\
96 & 96  & 192 & 27936 \\
96 & 168 & 336 & 73080 \\
96 & 360 & 720 & 294840 \\
\hline
\end{tabular}

\end{table}

\textbf{Evaluation metrics.} 
We report the MAE and MSE on test data where lower metrics indicate better results. We note that our model apply the same function for every channel of the test data (i.e. channel independence) and return the matrix output which contains $O$ time steps and $D$ features (the number of features of the corresponding testing data). 
Let $P \in \mathbb{R}^{D \times O}$ be the predicted value of our model and $V\in \mathbb{R}^{D \times O}$ be the ground truth value. The metrics are presented as follows: 
\begin{align*}
\text{MAE}(P, V) & = \frac{1}{D O}\sum_{d=1}^D \sum_{t=1}^O | P_t^d-   V_t^d|, \quad
\text{MSE}(P, V)  =\frac{1}{D O}\sum_{d=1}^D \sum_{t=1}^O ( P_t^d-   V_t^d)^2.
\end{align*}
In our test, we use the same metrics as the prior reference \citep{autoformer} with batch size 32. Note that we only implemented our results, the test results of other methods are from the TimesNet reference \citep{timesnet}.
\section{Experiment Results}


\renewcommand{\arraystretch}{1.2}
\begin{table}[ht!]
\caption{Comparisons of the test performance from our finetuned model and the variations. We report the average MSE and MAE over four prediction lengths: 96, 192, 336, 720.
}\label{tab:var2}
\centering
\begin{tabular}{|c|c|cc|cc|cc|cc|}
\hline
 & Models & \multicolumn{2}{c|}{Standard} & \multicolumn{2}{c|}{Variation 1} & \multicolumn{2}{c|}{Variation 2} & \multicolumn{2}{c|}{Variation 3} \\
 \cline{2-10}
\multirow{-2}{*}{Data} & Metric & MSE & MAE & MSE & MAE & MSE & MAE & MSE & MAE \\
\hline
\cellcolor[HTML]{FFF7E7} & 96 & {\color[HTML]{FF0000} \textbf{0.347}} & {\color[HTML]{FF0000} \textbf{0.373}} & {\color[HTML]{0000FF} \underline{0.354}} & {\color[HTML]{0000FF} \underline{0.393}} & 0.414 & 0.400 & 0.423 & 0.433 \\
\cellcolor[HTML]{FFF7E7} & 192 & {\color[HTML]{FF0000} \textbf{0.384}} & {\color[HTML]{FF0000} \textbf{0.393}} & {\color[HTML]{0000FF} \underline{0.389}} & {\color[HTML]{0000FF} \underline{0.409}} & 0.446 & 0.447 & 0.458 & 0.449 \\
\cellcolor[HTML]{FFF7E7} & 336 & {\color[HTML]{FF0000} \textbf{0.414}} & {\color[HTML]{FF0000} \textbf{0.415}} & {\color[HTML]{0000FF} \underline{0.418}} & {\color[HTML]{0000FF} \underline{0.429}} & 0.480 & 0.473 & 0.488 & 0.471 \\
\cellcolor[HTML]{FFF7E7} & 720 & {\color[HTML]{FF0000} \textbf{0.473}} & {\color[HTML]{FF0000} \textbf{0.451}} & {\color[HTML]{0000FF} \underline{0.479}} & {\color[HTML]{0000FF} \underline{0.464}} & 0.554 & 0.523 & 0.550 & 0.509 \\
\cline{2-10}\multirow{-5}{*}{\cellcolor[HTML]{FFF7E7}ETTm1} & Avg. & {\color[HTML]{FF0000} \textbf{0.405}} & {\color[HTML]{FF0000} \textbf{0.408}} & {\color[HTML]{0000FF} \underline{0.410}} & {\color[HTML]{0000FF} \underline{0.424}} & 0.473 & 0.461 & 0.480 & 0.465 \\
\hline
\cellcolor[HTML]{FFF7E7} & 96 & {\color[HTML]{FF0000} \textbf{0.385}} & {\color[HTML]{FF0000} \textbf{0.398}} & 0.428 & {\color[HTML]{0000FF} \underline{0.448}} & {\color[HTML]{0000FF} \underline{0.412}} & 0.484 & 0.456 & 0.481 \\
\cellcolor[HTML]{FFF7E7} & 192 & {\color[HTML]{FF0000} \textbf{0.432}} & {\color[HTML]{FF0000} \textbf{0.425}} & 0.473 & {\color[HTML]{0000FF} \underline{0.474}} & {\color[HTML]{0000FF} \underline{0.461}} & 0.512 & 0.504 & 0.510 \\
\cellcolor[HTML]{FFF7E7} & 336 & {\color[HTML]{FF0000} \textbf{0.473}} & {\color[HTML]{FF0000} \textbf{0.448}} & 0.512 & {\color[HTML]{0000FF} \underline{0.494}} & {\color[HTML]{0000FF} \underline{0.503}} & 0.537 & 0.546 & 0.533 \\
\cellcolor[HTML]{FFF7E7} & 720 & {\color[HTML]{FF0000} \textbf{0.492}} & {\color[HTML]{FF0000} \textbf{0.490}} & {\color[HTML]{0000FF} \underline{0.515}} & {\color[HTML]{0000FF} \underline{0.522}} & 0.527 & 0.580 & 0.566 & 0.576 \\
\cline{2-10}\multirow{-5}{*}{\cellcolor[HTML]{FFF7E7}ETTh1} & Avg. & {\color[HTML]{FF0000} \textbf{0.446}} & {\color[HTML]{FF0000} \textbf{0.440}} & 0.482 & {\color[HTML]{0000FF} \underline{0.484}} & {\color[HTML]{0000FF} \underline{0.476}} & 0.528 & 0.518 & 0.525 \\
\hline
\cellcolor[HTML]{FFF7E7} & 96 & {\color[HTML]{0000FF} \underline{0.081}} & {\color[HTML]{FF0000} \textbf{0.204}} & {\color[HTML]{FF0000} \textbf{0.080}} & {\color[HTML]{0000FF} \underline{0.220}} & 0.082 & 0.236 & 0.082 & 0.227 \\
\cellcolor[HTML]{FFF7E7} & 192 & {\color[HTML]{FF0000} \textbf{0.164}} & {\color[HTML]{FF0000} \textbf{0.300}} & {\color[HTML]{0000FF} \underline{0.165}} & {\color[HTML]{0000FF} \underline{0.318}} & 0.168 & 0.333 & 0.169 & 0.324 \\
\cellcolor[HTML]{FFF7E7} & 336 & {\color[HTML]{FF0000} \textbf{0.295}} & {\color[HTML]{FF0000} \textbf{0.407}} & {\color[HTML]{0000FF} \underline{0.296}} & {\color[HTML]{0000FF} \underline{0.434}} & 0.314 & 0.440 & 0.315 & 0.438 \\
\cellcolor[HTML]{FFF7E7} & 720 & {\color[HTML]{FF0000} \textbf{0.535}} & {\color[HTML]{FF0000} \textbf{0.587}} & {\color[HTML]{FF0000} \textbf{0.535}} & {\color[HTML]{0000FF} \underline{0.614}} & 0.561 & 0.646 & {\color[HTML]{0000FF} \underline{0.560}} & 0.703 \\
\cline{2-10}\multirow{-5}{*}{\cellcolor[HTML]{FFF7E7}Exchange} & Avg. & {\color[HTML]{FF0000} \textbf{0.269}} & {\color[HTML]{FF0000} \textbf{0.375}} & {\color[HTML]{FF0000} \textbf{0.269}} & {\color[HTML]{0000FF} \underline{0.396}} & {\color[HTML]{0000FF} \underline{0.281}} & 0.414 & 0.282 & 0.423 \\
\hline
\cellcolor[HTML]{FFF7E7} & 96 & {\color[HTML]{FF0000} \textbf{0.197}} & {\color[HTML]{FF0000} \textbf{0.282}} & 0.210 & {\color[HTML]{0000FF} \underline{0.294}} & {\color[HTML]{0000FF} \underline{0.204}} & 0.330 & 0.227 & 0.309 \\
\cellcolor[HTML]{FFF7E7} & 192 & {\color[HTML]{FF0000} \textbf{0.195}} & {\color[HTML]{FF0000} \textbf{0.282}} & 0.215 & {\color[HTML]{0000FF} \underline{0.299}} & {\color[HTML]{0000FF} \underline{0.206}} & 0.331 & 0.229 & 0.313 \\
\cellcolor[HTML]{FFF7E7} & 336 & {\color[HTML]{FF0000} \textbf{0.207}} & {\color[HTML]{FF0000} \textbf{0.296}} & 0.228 & {\color[HTML]{0000FF} \underline{0.313}} & {\color[HTML]{0000FF} \underline{0.220}} & 0.338 & 0.242 & 0.328 \\
\cellcolor[HTML]{FFF7E7} & 720 & {\color[HTML]{FF0000} \textbf{0.242}} & {\color[HTML]{FF0000} \textbf{0.229}} & 0.256 & {\color[HTML]{0000FF} \underline{0.288}} & {\color[HTML]{0000FF} \underline{0.255}} & 0.352 & 0.274 & 0.331 \\
\cline{2-10}\multirow{-5}{*}{\cellcolor[HTML]{FFF7E7}Electricity} & Avg. & {\color[HTML]{FF0000} \textbf{0.210}} & {\color[HTML]{FF0000} \textbf{0.272}} & 0.227 & {\color[HTML]{0000FF} \underline{0.298}} & {\color[HTML]{0000FF} \underline{0.221}} & 0.338 & 0.243 & 0.320 \\
\hline
\cellcolor[HTML]{F1E5FF} & 96 & {\color[HTML]{FF0000} \textbf{0.196}} & {\color[HTML]{FF0000} \textbf{0.294}} & 0.237 & 0.342 & {\color[HTML]{0000FF} \underline{0.200}} & {\color[HTML]{0000FF} \underline{0.310}} & 0.257 & 0.359 \\
\cellcolor[HTML]{F1E5FF} & 192 & {\color[HTML]{FF0000} \textbf{0.266}} & {\color[HTML]{FF0000} \textbf{0.342}} & 0.304 & 0.386 & {\color[HTML]{0000FF} \underline{0.286}} & {\color[HTML]{0000FF} \underline{0.371}} & 0.336 & 0.416 \\
\cellcolor[HTML]{F1E5FF} & 336 & {\color[HTML]{FF0000} \textbf{0.365}} & {\color[HTML]{FF0000} \textbf{0.412}} & {\color[HTML]{0000FF} \underline{0.384}} & {\color[HTML]{0000FF} \underline{0.440}} & 0.401 & 0.448 & 0.419 & 0.475 \\
\cellcolor[HTML]{F1E5FF} & 720 & {\color[HTML]{FF0000} \textbf{0.492}} & {\color[HTML]{FF0000} \textbf{0.485}} & {\color[HTML]{0000FF} \underline{0.498}} & {\color[HTML]{0000FF} \underline{0.504}} & 0.581 & 0.545 & 0.568 & 0.555 \\
\cline{2-10}\multirow{-5}{*}{\cellcolor[HTML]{F1E5FF}ETTm2} & Avg. & {\color[HTML]{FF0000} \textbf{0.330}} & {\color[HTML]{FF0000} \textbf{0.383}} & {\color[HTML]{0000FF} \underline{0.356}} & {\color[HTML]{0000FF} \underline{0.418}} & 0.367 & {\color[HTML]{0000FF} \underline{0.418}} & 0.395 & 0.451 \\
\hline
\cellcolor[HTML]{F1E5FF} & 96 & {\color[HTML]{FF0000} \textbf{0.316}} & {\color[HTML]{FF0000} \textbf{0.372}} & 0.350 & {\color[HTML]{0000FF} \underline{0.417}} & 0.340 & 0.450 & {\color[HTML]{0000FF} \underline{0.333}} & 0.440 \\
\cellcolor[HTML]{F1E5FF} & 192 & {\color[HTML]{FF0000} \textbf{0.419}} & {\color[HTML]{FF0000} \textbf{0.433}} & {\color[HTML]{0000FF} \underline{0.442}} & {\color[HTML]{0000FF} \underline{0.471}} & 0.453 & 0.514 & 0.453 & 0.512 \\
\cellcolor[HTML]{F1E5FF} & 336 & {\color[HTML]{FF0000} \textbf{0.536}} & {\color[HTML]{FF0000} \textbf{0.507}} & {\color[HTML]{0000FF} \underline{0.537}} & {\color[HTML]{0000FF} \underline{0.528}} & 0.565 & 0.578 & 0.554 & 0.573 \\
\cellcolor[HTML]{F1E5FF} & 720 & {\color[HTML]{FF0000} \textbf{0.708}} & {\color[HTML]{FF0000} \textbf{0.543}} & {\color[HTML]{0000FF} \underline{0.725}} & {\color[HTML]{0000FF} \underline{0.575}} & 0.769 & 0.653 & 0.748 & 0.644 \\
\cline{2-10}\multirow{-5}{*}{\cellcolor[HTML]{F1E5FF}ETTh2} & Avg. & {\color[HTML]{FF0000} \textbf{0.495}} & {\color[HTML]{FF0000} \textbf{0.464}} & {\color[HTML]{0000FF} \underline{0.514}} & {\color[HTML]{0000FF} \underline{0.498}} & 0.532 & 0.549 & 0.522 & 0.542 \\
\hline
\cellcolor[HTML]{F1E5FF} & 96 & {\color[HTML]{FF0000} \textbf{0.664}} & {\color[HTML]{FF0000} \textbf{0.410}} & {\color[HTML]{0000FF} \underline{0.665}} & {\color[HTML]{0000FF} \underline{0.419}} & 0.706 & 0.440 & 0.791 & 0.421 \\
\cellcolor[HTML]{F1E5FF} & 192 & {\color[HTML]{FF0000} \textbf{0.606}} & {\color[HTML]{FF0000} \textbf{0.380}} & {\color[HTML]{0000FF} \underline{0.624}} & {\color[HTML]{0000FF} \underline{0.408}} & 0.670 & 0.421 & 0.754 & 0.423 \\
\cellcolor[HTML]{F1E5FF} & 336 & {\color[HTML]{FF0000} \textbf{0.608}} & {\color[HTML]{FF0000} \textbf{0.378}} & {\color[HTML]{0000FF} \underline{0.631}} & {\color[HTML]{0000FF} \underline{0.405}} & 0.677 & 0.422 & 0.758 & 0.429 \\
\cellcolor[HTML]{F1E5FF} & 720 & {\color[HTML]{FF0000} \textbf{0.647}} & {\color[HTML]{FF0000} \textbf{0.398}} & {\color[HTML]{0000FF} \underline{0.651}} & {\color[HTML]{0000FF} \underline{0.415}} & 0.719 & 0.442 & 0.790 & 0.439 \\
\cline{2-10}\multirow{-5}{*}{\cellcolor[HTML]{F1E5FF}Traffic} & Avg. & {\color[HTML]{FF0000} \textbf{0.631}} & {\color[HTML]{FF0000} \textbf{0.392}} & {\color[HTML]{0000FF} \underline{0.643}} & {\color[HTML]{0000FF} \underline{0.412}} & 0.693 & 0.431 & 0.773 & 0.428 \\
\hline
\cellcolor[HTML]{F1E5FF} & 96 & {\color[HTML]{FF0000} \textbf{0.194}} & {\color[HTML]{FF0000} \textbf{0.251}} & 0.210 & 0.304 & {\color[HTML]{0000FF} \underline{0.197}} & {\color[HTML]{0000FF} \underline{0.301}} & 0.239 & 0.302 \\
\cellcolor[HTML]{F1E5FF} & 192 & {\color[HTML]{FF0000} \textbf{0.235}} & {\color[HTML]{FF0000} \textbf{0.291}} & 0.251 & {\color[HTML]{0000FF} \underline{0.333}} & {\color[HTML]{0000FF} \underline{0.240}} & 0.346 & 0.280 & 0.340 \\
\cellcolor[HTML]{F1E5FF} & 336 & {\color[HTML]{FF0000} \textbf{0.281}} & {\color[HTML]{FF0000} \textbf{0.327}} & 0.304 & {\color[HTML]{0000FF} \underline{0.368}} & {\color[HTML]{0000FF} \underline{0.290}} & 0.378 & 0.330 & 0.379 \\
\cellcolor[HTML]{F1E5FF} & 720 & {\color[HTML]{FF0000} \textbf{0.344}} & {\color[HTML]{FF0000} \textbf{0.368}} & 0.363 & {\color[HTML]{0000FF} \underline{0.370}} & {\color[HTML]{0000FF} \underline{0.357}} & 0.446 & 0.390 & 0.396 \\
\cline{2-10}\multirow{-5}{*}{\cellcolor[HTML]{F1E5FF}Weather} & Avg. & {\color[HTML]{FF0000} \textbf{0.264}} & {\color[HTML]{FF0000} \textbf{0.309}} & 0.282 & {\color[HTML]{0000FF} \underline{0.344}} & {\color[HTML]{0000FF} \underline{0.271}} & 0.368 & 0.310 & 0.354 \\
\hline
\end{tabular}
\end{table}
\subsection{Variations Analysis}

In this section, we analyze some variations of our model. In the first variation, instead of applying the contrastive loss directly on the representation $z$, we use another decoder to transform it to a representation $y$, then apply the contrastive learning on $y$. We also use a linear layer for the decoder transforming $z$ to $y$. We choose the same dimension for $z$ as the original approach (as described in Table \ref{tab:model}). For the dimension of $y$, we do grid search between three choices (half the dimention of $z$, same dimension of $z$ and double the dimension of $z$) and report the best results of these choices. 

For the second variation, we do not use a decoder for the prediction output (i.e. using an identity layer in Figure \ref{fig:des}(a) instead of the decoder). In the final variation, we replace the linear encoder and decoder by two-layer neural networks. We keep the same representation dimension for $z$ as described in Table \ref{tab:model}. We also perform grid search on the number of hidden layers to find the best (training) model, then report the test result of that model.

We report the full results in Table \ref{tab:var2}. This analysis shows that the linear encoder-decoder is essential for the good generalization performance of our method. The first variation with two decoders is also able to capture the dynamic, its performance has $6.2\%$ difference in average compared to the implementation of contrastive learning directly on the representation $z$. The second variation shows that a decoder for the prediction output is needed. 
\subsection{Parameters Analysis}
In this section, we analyze how the performance of the model changes when the model parameters $\lambda$ changes. Note that $\lambda$ and $\tau$ are the regularization factor and the temperature parameter of the model, respectively, which controls the supervised contrastive loss term. 

\subsubsection{Parameter $\lambda$}
\begin{figure}[ht]
    \centering
    \includegraphics[width=0.85\textwidth]{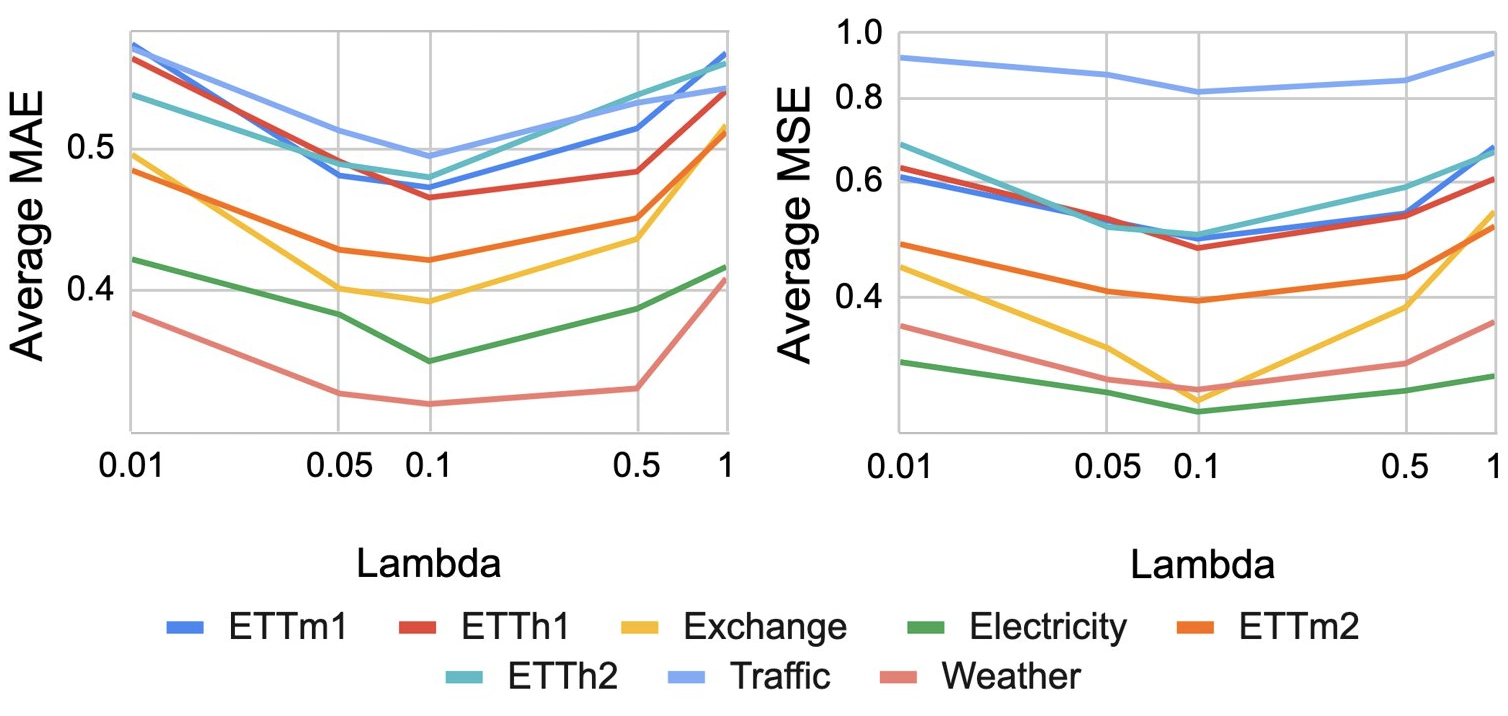}
    \caption{The average test performance of the pretrained model by parameter $\lambda$, for eights datasets. We plot both the $x$-axis and $y$-axis in logarithm scale to highlight the differences in accuracy. } 
    \label{fig:lamb2}
\end{figure}

Here we set the temperature parameter of the model $\tau$ to be 0.1.
The choice of the regularization parameter $\lambda$ varies in $[0.01, 0.05, 0.1, 0.5, 1]$. We perform the test for eight datasets using the pretrained models and report the average errors (MAE and MSE) over the four prediction lengths in Table \ref{tab:lamb} and Figure \ref{fig:lamb2}. We observe that in most datasets, the choice $\lambda = 0.1$ performs the best. 

\setlength\doublerulesep{3pt}
\setlength{\tabcolsep}{8pt}
\begin{table}[h]
\caption{The average test performance of the pretrained model by parameter $\lambda$, for eights datasets. We report the average MSE and MAE over four prediction lengths: 96, 192, 336, 720.
}\label{tab:lamb}
\centering
\begin{tabular}{|c|c|c|c|c|c|c|c|c|}
\hline
\multicolumn{9}{|c|}{MSE} \\
\hline
$\lambda$ & ETTm1 & ETTh1 & Exchange & Electricity & ETTm2 & ETTh2 & Traffic & Weather \\
\hline
0.01 & 0.608 & 0.628 & 0.445 & 0.320 & 0.482 & 0.681 & 0.920 & 0.363 \\
0.05 & 0.523 & 0.526 & 0.336 & 0.288 & 0.409 & 0.511 & 0.867 & 0.301 \\
0.1 & 0.491 & 0.475 & 0.280 & 0.270 & 0.396 & 0.498 & 0.817 & 0.291 \\
0.5 & 0.536 & 0.531 & 0.387 & 0.290 & 0.430 & 0.587 & 0.850 & 0.319 \\
1 & 0.676 & 0.605 & 0.539 & 0.305 & 0.513 & 0.663 & 0.935 & 0.368 \\
\hline
\hline
\multicolumn{9}{|c|}{MAE} \\
\hline
$\lambda$ & ETTm1 & ETTh1 & Exchange & Electricity & ETTm2 & ETTh2 & Traffic & Weather \\
\hline
0.01 & 0.589 & 0.576 & 0.495 & 0.420 & 0.483 & 0.544 & 0.585 & 0.386 \\
0.05 & 0.479 & 0.490 & 0.401 & 0.385 & 0.426 & 0.488 & 0.514 & 0.340 \\
0.1 & 0.470 & 0.463 & 0.393 & 0.358 & 0.420 & 0.478 & 0.494 & 0.335 \\
0.5 & 0.516 & 0.482 & 0.434 & 0.389 & 0.448 & 0.544 & 0.537 & 0.343 \\
1 & 0.581 & 0.548 & 0.519 & 0.415 & 0.513 & 0.572 & 0.550 & 0.408 \\
\hline
\end{tabular}
\end{table}

\subsubsection{Parameter $\tau$}
\begin{figure}[ht]
    \centering
    \includegraphics[width=0.85\textwidth]{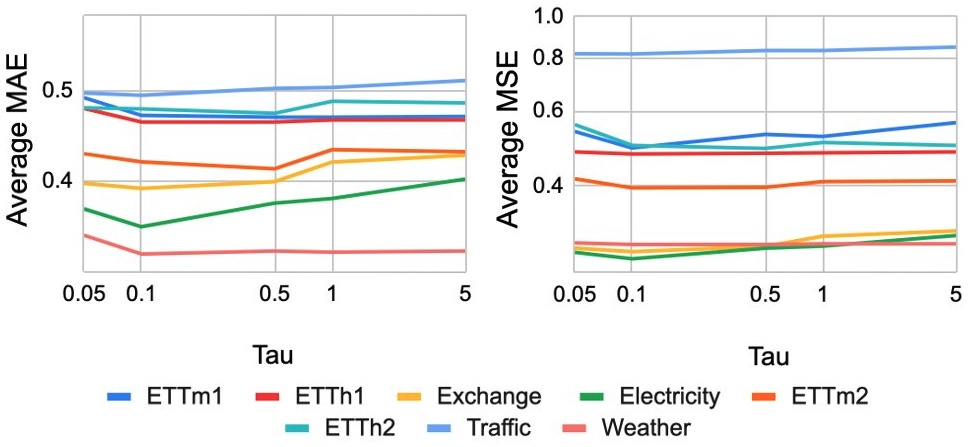}
    \caption{The average test performance of the pretrained model by parameter $\tau$, for eights datasets. We plot both the $x$-axis and $y$-axis in logarithm scale to highlight the differences in accuracy. } 
    \label{fig:tau2}
\end{figure}

We set the regularization factor $\lambda$ to be 0.1.
The choice of the temperature parameter $\tau$ varies in $[0.05, 0.1, 0.5, 1, 5]$. We perform the test for eight datasets using the pretrained models and report the average errors (MAE and MSE) over the four prediction lengths in Table \ref{tab:tau} and Figure \ref{fig:tau2}. Since the choice $\tau = 0.1$ performs the reasonably well in most of the datasets, we choose $\tau = 0.1$ in our experiment. 

\setlength\doublerulesep{3pt}
\setlength{\tabcolsep}{8pt}
\begin{table}[h]
\caption{The average test performance of the pretrained model by parameter $\tau$, for eights datasets. We report the average MSE and MAE over four prediction lengths: 96, 192, 336, 720.
}\label{tab:tau}
\centering
\begin{tabular}{|c|c|c|c|c|c|c|c|c|}
\hline
\multicolumn{9}{|c|}{MSE} \\
\hline
$\tau$ & ETTm1 & ETTh1 & Exchange & Electricity & ETTm2 & ETTh2 & Traffic & Weather \\
\hline
0.05& 0.537& 0.481& 0.285& 0.279& 0.416& 0.558& 0.819& 0.294\\
0.1& 0.491& 0.475& 0.280& 0.270& 0.396& 0.498& 0.817& 0.291\\
0.5& 0.529& 0.478& 0.288& 0.285& 0.397& 0.490& 0.833& 0.291\\
1& 0.523& 0.479& 0.304& 0.289& 0.410& 0.506& 0.833& 0.292\\
5& 0.563& 0.481& 0.313& 0.306& 0.411& 0.498& 0.849& 0.292\\
\hline
\hline
\multicolumn{9}{|c|}{MAE} \\
\hline
$\tau$ & ETTm1 & ETTh1 & Exchange & Electricity & ETTm2 & ETTh2 & Traffic & Weather \\
\hline
0.05& 0.491& 0.478& 0.398& 0.374& 0.428& 0.479& 0.497& 0.350\\
0.1& 0.470& 0.463& 0.393& 0.358& 0.420& 0.478& 0.494& 0.335\\
0.5& 0.468& 0.463& 0.400& 0.379& 0.412& 0.473& 0.503& 0.337\\
1& 0.468& 0.465& 0.419& 0.383& 0.432& 0.487& 0.504& 0.336\\
5& 0.469& 0.465& 0.426& 0.402& 0.430& 0.485& 0.512& 0.337\\
\hline
\end{tabular}
\end{table}

\newpage

\bibliographystyle{plainnat}
\bibliography{refs}





\end{document}